\pdfoutput=1
\documentclass[11pt]{article}

\usepackage[]{EMNLP2022}

\usepackage{times}
\usepackage{latexsym}
\usepackage{caption}
\usepackage{subcaption}
\usepackage{graphicx}
\usepackage{subfloat}
\usepackage{booktabs} % To thicken table lines
\usepackage{multirow}
\usepackage{amsmath,amssymb}

\usepackage[T1]{fontenc}
\usepackage[utf8]{inputenc}

\usepackage{microtype}

\usepackage{inconsolata}

\usepackage{xcolor}

\title{On the Calibration of Massively Multilingual Language Models}

\author{Kabir Ahuja\textsuperscript{1} \quad  Sunayana Sitaram\textsuperscript{1} \quad Sandipan Dandapat\textsuperscript{2}  \quad Monojit Choudhury\textsuperscript{2}\\
\textsuperscript{1} Microsoft Research, India \\
\textsuperscript{2} Microsoft R\&D, India \\
{\tt \small \{t-kabirahuja,sadandap,sunayana.sitaram,monojitc\}@microsoft.com}
}

\begin{document}
\maketitle
\begin{abstract}
Massively Multilingual Language Models (MMLMs) have recently gained popularity due to their surprising effectiveness in cross-lingual transfer. While there has been much work in evaluating these models for their performance on a variety of tasks and languages, little attention has been paid on how well \textit{calibrated} these models are with respect to the confidence in their predictions. We first investigate the calibration of MMLMs in the zero-shot setting and observe a clear case of miscalibration in low-resource languages or those which are typologically diverse from English. Next, we empirically show that calibration methods like temperature scaling and label smoothing do reasonably well towards improving calibration in the zero-shot scenario. We also find that few-shot examples in the language can further help reduce the calibration errors, often substantially. Overall, our work contributes towards building more reliable multilingual models by highlighting the issue of their miscalibration, understanding what language and model specific factors influence it, and pointing out the strategies to improve the same.

% Next, we show that standard calibration methods like Temperature Scaling and Label Smoothing can be used to substantially improve calibration in the zero-shot scenario which can be even further improved by collecting few-shot examples in those languages. Overall, our work provides a step towards building more reliable multilingual models by taking into account their calibration in addition to their performance across languages. %Overall, our work contributes towards understanding and building more reliable multilingual models by highlighting calibration as an issue in these models, which should be considered along with performance while deploying them into production.

% Overall, our work provides a step towards building more reliable multilingual models by taking into account their calibration in addition to their performance across languages.

%Overall, our work provides a step towards building more reliable multilingual models and draw attention of the community to consider calibration of these models along with their performance while deploying them into production

\end{abstract}

\section{Introduction}

\begin{figure*}
    \centering
    % \captionsetup[subfloat]{margin=0.5em}
    \begin{subfigure}{.24\textwidth}
    \centering
    \captionsetup{justification=centering}
    \includegraphics[width=.99\textwidth]{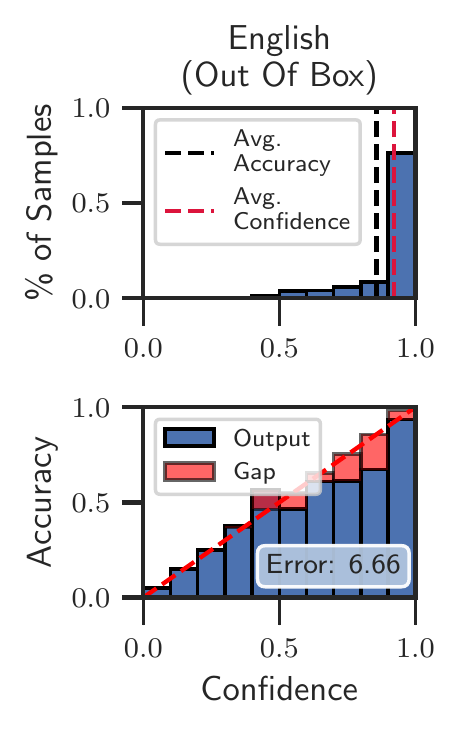}
    \vspace*{-7mm}
    \caption{}%Out of box calibration on English}
    \label{fig:en_oob}
    \end{subfigure}
    \begin{subfigure}{.24\textwidth}
    \centering
    \includegraphics[width=.99\textwidth]{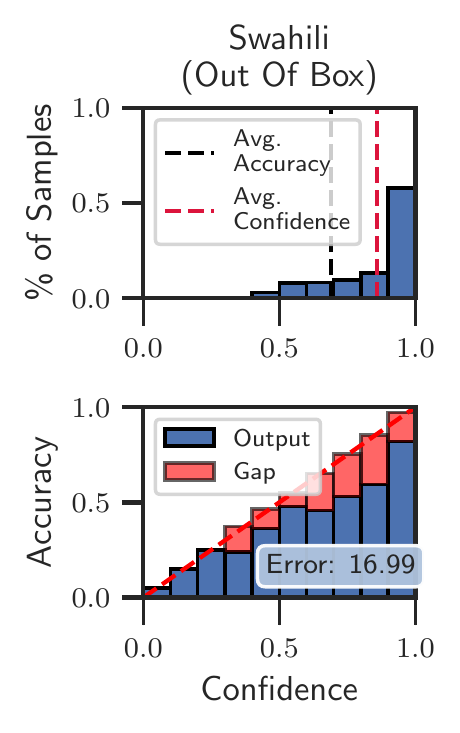}
    \vspace*{-7mm}
    \caption{}%Out of box calibration on Swahili}
    \label{fig:sw_oob}
    \end{subfigure}
    \begin{subfigure}{.24\textwidth}
    \centering
    \includegraphics[width=.99\textwidth]{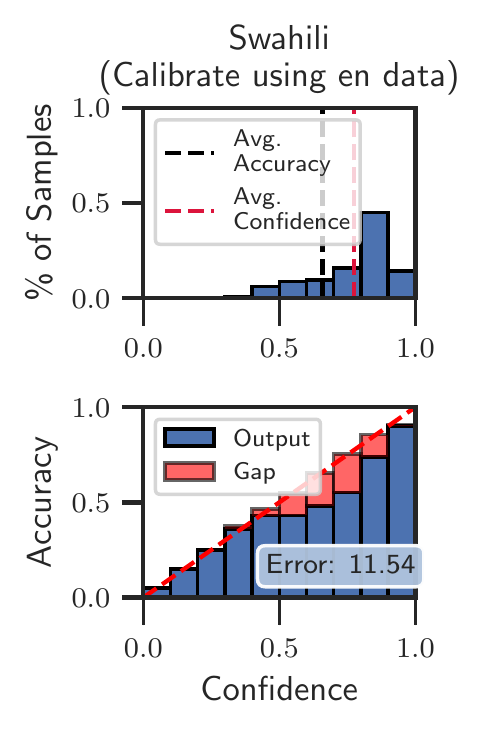}
    \vspace*{-7mm}
    \caption{}%Calibrating using zero-shot methods}
    \label{fig:sw_zs}
    \end{subfigure}
    \begin{subfigure}{.24\textwidth}
    \centering
    \includegraphics[width=.99\textwidth]{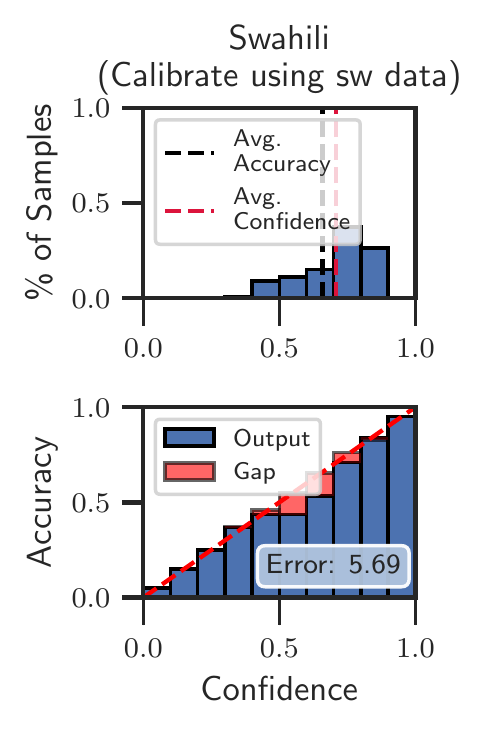}
    \vspace*{-7mm}
    \caption{}%Calibrating using few-shot methods}
    \label{fig:sw_fs}
    \end{subfigure}
    \caption{Reliability Diagrams for XLMR fine-tuned on XNLI using training data in English. 
    % For a perfectly calibrated model confidences should match accuracies i.e. the blue bars should lie close to the red-dotted line and the gap between the two represents calibration error.
    \ref{fig:en_oob} and \ref{fig:sw_oob} provides the calibration of XLMR fine-tuned without any calibration technique on English and Swahilli. \ref{fig:sw_zs} and \ref{fig:sw_fs} are obtained by calibrating the model using TS + LS and Self-TS + LS techniques respectively as described in Section \ref{sec:cal_methods}}
    \label{fig:cal_en_sw}
\end{figure*}

Massively Multilingual Language Models (MMLMs) like mBERT \cite{devlin-etal-2019-bert}, XLMR \cite{conneau-etal-2020-unsupervised}, mT5 \cite{xue-etal-2021-mt5} and mBART \cite{liu-etal-2020-multilingual-denoising} have been surprisingly effective at zero-shot cross lingual transfer i.e. when fine-tuned on an NLP task in one language, they often tend to generalize reasonably well in languages unseen during fine-tuning. 

These models have been evaluated for their performance across a range of multilingual tasks \cite{Pan2017, nivre2018universal, Conneau2018xnli} and numerous methods like adapters \cite{pfeiffer-etal-2020-mad}, sparse fine-tuning \cite{ansell-etal-2022-composable} and few-shot learning \cite{lauscher-etal-2020-zero} have been proposed to further improve performance of cross lingual transfer. 

Despite these developments, there has been little to no attention paid to the calibration of these models across languages i.e. how reliable  the confidence predictions of these models are. 
As these models find their way more and more into the real word applications with safety implications, like Hate Speech Detection \cite{DavidsonWMW17, deshpande-2022-highly} it becomes important to only take extreme actions for high confidence predictions by the model \cite{Sarkar_KhudaBukhsh_2021}. Hence, calibrated confidences are desirable to have when deploying such systems in practice. 

\citet{guo-2017-on} showed that modern neural networks used for Image Recognition \cite{he-et-al-2016-deep} perform much better than the ones introduced decades ago \cite{lecun-et-al-1998-gradient}, but are significantly worse calibrated and often over-estimate their confidence on incorrect predictions. For NLP tasks specifically, \citet{desai-durrett-2020-calibration} showed that classifiers trained using pre-trained transformer based models \cite{devlin-etal-2019-bert} are well calibrated both in-domain and out-of-domain settings compared to non-pre-trained model baselines \cite{chen-etal-2017-enhanced}. Notably, \citet{ponti-etal-2021-parameter} highlights, since zero-shot cross lingual transfer represents shifts in the data distribution the point estimates are likely to be miscalibrated, which forms the core setting of our work.
% Aside from classification, calibration in other NLP tasks like translation \cite{aviral-sarawagi-2019-calibration, wang-etal-2020-inference} and language modelling \cite{braverman2020calibration, Zhao-et-al-2021-calibrate}, but for the scope of our work we focus on classification tasks and follow the setup of \citet{desai-durrett-2020-calibration}.

% In context of Neural Machine Translation \citet{aviral-sarawagi-2019-calibration} and \citet{wang-etal-2020-inference} also observed a case of mis-calibration with the latter finding correlations between the calibration errors and certain linguistic properties. Recently, \citet{Zhao-et-al-2021-calibrate} showed calibrating pre-trained langauge models like GPT-3 can substantially improve their accuracy as well as stability across different prompt choices. For the scope of our work we focus on classification tasks and follow the setup of \citet{desai-durrett-2020-calibration}.

In light of this, our work has three main contributions. \textit{First}, we investigate the calibration of two commonly used MMLMs: mBERT and XLM-R on four NLU tasks under zero-shot setting where the models are fine-tuned in English and calibration errors are computed on unseen languages. We find a clear increase in calibration errors compared to English as can be seen in Figures \ref{fig:en_oob} and \ref{fig:sw_oob}, with calibration being significantly worse for Swahili compared to English.
% the average confidence being substantially higher than the average accuracy for Swahili compared to English.

\textit{Second}, we look for factors that might affect the zero-shot calibration of MMLMs and find in most cases that the calibration error is strongly correlated with pre-training data size, syntactic similarity and sub-word overlap between the unseen language and English. This reveals that MMLMs are mis-calibrated in the zero-shot setting for low-resource languages and the languages that are typologically distant from English. 

\textit{Finally}, we show that model calibration across different languages can be substantially improved by utilizing standard calibration techniques like Temperature Scaling \cite{guo-2017-on} and Label Smoothing \cite{pereyra-et-al-2017} without collecting any data in the language (see Figure \ref{fig:sw_zs}). Using a few examples in a language (the few-shot setting), we see even more significant drops in the calibration errors as can be seen in Figure \ref{fig:sw_fs}.

To the best of our knowledge, ours is the first work to investigate and improve the calibration of MMLMs. We expect this study to be a significant contribution towards building reliable and linguistically fair multilingual models. To encourage future research in the area we make our code publicly available\footnote{\url{https://github.com/microsoft/MMLMCalibration}}.

%, and to ensure future research in the area we will make our code public.

\section{Calibration of Pre-trained MMLMs}

Consider a classifier $h :  \mathcal{X} \rightarrow [K]$ obtained by fine-tuning an MMLM for some task with training data in a pivot language $p$, where $[K]$ denotes the set of labels $\{1, 2, \cdots, K\}$. We assume $h$ can predict confidence of each of the $[K]$ labels and is given by $h_{k}(x) \in [0,1]$ for the $k^{th}$ label. 
% Consider an MMLM $\mathcal{M}$ pre-trained on a set of languages given by $\mathcal{L}$. We fine-tune $\mathcal{M}$ on a multi-class classification task $\mathfrak{T}$ using labelled data $(\mathcal{X}_p, \mathcal{Y}_p)$ in a language $p \in \mathcal{L}$ to obtain a classifier $h : \mathcal{X} \rightarrow [K]$, where $[K]$ denotes the set of labels $\{1, 2, \cdots, K\}$. We assume $h$ can predict confidence of each of the $[K]$ labels and is given by $h_{k}(x) \in [0,1]$ for the $k^{th}$ label. 
$h$ is said to be calibrated if the following equality holds:
\begin{equation*}
    p (y = k | h_k(x) = q) = q
\end{equation*}
In other words, for a perfectly calibrated classifier, if the predicted confidence for a label $k$ on an input $x$ is $q$, then with a probability $q$ the input should actually be labelled $k$. Naturally, in practical settings the equality does not hold, and neural network based classifiers are often miscalibrated \cite{guo-2017-on}.
%found to over-estimate the confidence in their predictions \cite{guo-2017-on}.% and in some cases under-estimate the confidence \cite{wang-etal-2020-inference} (LHS > RHS). 
 One way of quantifying this notion of miscalibration is through the Expected Calibration Error (ECE) which is defined as the difference in expectation between the confidence in classifier's predictions and their accuracies (Refer Appendix \ref{sec:cal_measure} for details).
%  in practice is computed by dividing the classifier's predictions into $M$ equally spaced bins and measuring the difference between the confidence and accuracy values for each bin as described in \citet{guo-2017-on}.
%  defined as expected value of the difference between the confidence and accuracy of the classifier's predictions. In practice ECE of a classifier is computed by dividing the predictions into $M$ equally spaced bins and measuring the difference as described in \citet{guo-2017-on} (Refer Appendix \myworries{add section} for details). 
%  where the confidence predictions on the $n$ test examples are grouped into $M$ uniform-sized bins, such that set of examples belonging to the $m^{th}$ bin are denoted by $B_m$. Accuracy ($\text{acc}(B_m)$) and average confidence ($\text{conf}(B_m)$) for each bin is computed and a weighted average of the differences between the two is taken to obtain ECE.
% \begin{equation}
%     \texttt{ECE} =  \sum_{m = 1}^M \frac{|B_m|}{n} |\text{acc}(B_m) - \text{conf}(B_m)|
% \end{equation}
In our experiments we compute ECE on each language $l$'s test data and denote their corresponding calibration errors as $\texttt{ECE}(l)$. 
% For zero-shot setting, we consider $l \in {\mathcal{L} - \{p\}}$, i.e. all languages other than the one used during fine-tuning which in most of our experiments is English.

\subsection{Calibration Methods}
\label{sec:cal_methods}
We briefly review some commonly used methods for calibrating neural network based classifiers.
% and the ones that we use in our experiments.

\noindent\textbf{1. Temperature Scaling (TS and Self-TS)}\cite{guo-2017-on}
% is a simple extension of Platt Scaling \cite{Platt99probabilisticoutputs} to multi-class classification that has been shown to greatly improve the calibration of neural network based classifiers \cite{guo-2017-on}. Temperature Scaling 
is applied by scaling the output logits using a temperature parameter $T$ before applying softmax i.e. :
\begin{equation*}
    h_k(x) = \frac{\exp{o_k(x)/T}}{\sum_{k' \in K}\exp{o_{k'}(x)/T}}
\end{equation*}
, where $o_k$ denotes the logits corresponding to the $k^{th}$ class. $T$ is a learnable parameter obtained post-training by maximizing the log-likelihood on the dev set while keeping other network parameters fixed. We experiment with two settings for improving calibration on a target language: using dev data in English to perform temperature scaling (TS) and using the target language's dev data (Self-TS). 
% Note that temperature scaling does not affect the accuracy of the original classifier as all the logits are scaled by the same factor. 
% and hence the maximum element essentially remains the same. 
% Although, conceptually simple this method been shown to greatly improve the calibration of neural network based classifiers \cite{guo-2017-on}.

% be  problems where the output logits are inversely scaled by a temperature parameter $T$, which is often a learnable parameter trained by maximizing the likelihood on the dev set. Although, conceptually simple this method been shown to greatly improve the calibration of neural network based classifiers \cite{guo-2017-on}. Temperature scaling also does not affect the accuracy of the original classifier as all the logits are scaled by the same factor and hence the maximum element essentially remains the same. In our experiments we use English dev dataset to learn the value of $T$ and call this method TS. We also consider using few-shot data in the target language for calibration using this method and denote that baseline as Self-TS.

\noindent \textbf{2. Label Smoothing (LS)} \cite{pereyra-et-al-2017} is a regularization technique that penalizes low entropy distributions by using \textit{soft} labels that are obtained by assigning a fixed probability $q = 1 - \alpha$ to the true label ($0< \alpha < 1$), and distributing the remaining probability mass uniformly across the remaining classes. %While this method was not specifically introduced for improving calibration,
Label smoothing has been empirically shown to be competitive with temperature scaling for calibration \cite{rafael-et-al-2019-when} especially in out of domain settings \cite{desai-durrett-2020-calibration}
% \footnote{Note that unlike temperature scaling, label smoothing can affect the accuracy of the original classifier.}.
% \citet{rafael-et-al-2019-when} showed empirically that training with label smoothing implicitly calibrates the classifiers and often performs competitively with Temperature Scaling. 
% \citet{desai-durrett-2020-calibration} showed that label smoothing is more effective for out-of-domain calibration than temperature smoothing for pre-trained transformer based models, which is particularly of interest to us, since we are dealing with zero-shot calibration on unseen languages.\footnote{Note that unlike temperature scaling, Label smoothing is applied during the training and hence can affect the accuracy of the original classifier.}

\noindent \textbf{3. Few-Shot Learning (FSL)} We also investigate if fine-tuning the MMLM on a few examples in a target language in addition to the data in English, leads to any improvement in calibration as it does in the performance \cite{lauscher-etal-2020-zero}. Since  these models are expected to be calibrated worse for out-of-domain data compared to in-domain data \cite{desai-durrett-2020-calibration}, we try to improve calibration by reducing the domain shift through few-shot learning.

Apart from these, we also consider combinations of different calibration methods in our experiments, including Label Smoothing with Temperature Scaling (TS + LS or Self-TS + LS) and Few-Shot Learning with Label Smoothing (FSL + LS).
\section{Experiments}
We seek to answer the following research questions: a) How well calibrated are fine-tuned MMLMs in the zero-shot cross lingual setting? b) What linguistic and model-specific factors influence calibration errors across languages? c) Can we improve the calibration of fine-tuned models across languages?
% We start by describing our experimental setup including the datasets and pre-trained MMLMs that we use for our experiments and then discuss the results to address the above questions.

\begin{table}[]
\centering
\small
\begin{tabular}{p{0.9cm}p{1.1cm}p{0.9cm}p{1.2cm}p{1.4cm}}
    \toprule
    Dataset & MMLM & $\texttt{ECE}(en)$ & $\displaystyle \mathop{\mathbb{E}}_{l \in \mathcal{L}'}[\texttt{ECE}(l)]$ & $\displaystyle \max_{l \in \mathcal{L}'}\texttt{ECE}(l)$\\
    \midrule
    \multirow{ 2}{*}{XNLI} & XLM-R & 7.32 & 13.34& 19.07 (sw)\\
    & mBERT & 5.44 & 12.34 & 45.15 (th) \\
    \midrule
    \multirow{ 2}{*}{XCOPA} & XLM-R & 14.54 & 20.07 & 29.33 (sw) \\
    & mBERT & 23.4 & 23.51 & 29.02 (sw)\\
    \midrule
    \multirow{ 2}{*}{MARC} & XLM-R & 7.15 & 9.65 & 13.45 (zh) \\
    & mBERT & 9.38 & 11.11 & 17.33 (ja)\\
    \midrule
    \multirow{ 2}{*}{PAWS-X} & XLM-R & 1.93 & 4.28 & 5.88 (ja) \\
    & mBERT & 3.57 & 10.32 & 15.65 (ko)\\
    \bottomrule

\end{tabular}
\caption{Calibration Errors across tasks for XLM-R and mBERT. $\mathcal{L}'$ in the fourth column denotes the set of supported languages in a task other than English. The language in parenthesis in the column 5 denotes the language with maximum calibration error.}
\label{tab:oob_cali}
\end{table}

\begin{table}[]
    \centering
    \small
    \begin{tabular}{p{1.5cm}p{1cm}p{1cm}p{1cm}}
        \toprule
         Dataset & SIZE & SYN & SWO \\
         \midrule
         XNLI & -0.8 & \textbf{-0.88} & -0.85 \\
         XCOPA & \textbf{-0.85} & -0.73 & -0.62 \\
         MARC & -0.41 & \textbf{-0.46} & -0.27 \\
         PAWS-X & -0.48 & \textbf{-0.93} & -0.92 \\
         \bottomrule
    \end{tabular}
    \caption{Pearson correlation coefficient of ECE with SIZE, SYN, and SWO features of different languages in the test set for XLMR.}
    \label{tab:cal_factors}
\end{table}

\begin{table*}[]
\centering
\small
\begin{tabular}{p{1cm}p{1.2cm}p{0.9cm}p{0.9cm}p{0.9cm}p{1.2cm}p{1.2cm}p{1.7cm}p{0.9cm}p{1.2cm}}
    \toprule
    \multirow{ 2}{*}{Dataset} & \multirow{ 2}{*}{MMLM} & \multicolumn{4}{c}{Zero-Shot Calibration} & \multicolumn{4}{c}{Few-Shot Calibration}\\
    \cmidrule(l){3-6}  \cmidrule(l){7-10}
    & & OOB & TS & LS & TS + LS & Self-TS & Self-TS + LS & FSL & FSL + LS\\
    \cmidrule(l){3-10}
    \multirow{ 2}{*}{XNLI} & XLM-R & 13.34 & 6.74 &  6.93 & \textbf{4.89}$^{\dagger}$ & 5.41 & \textbf{4.05}$^{\ddag}$ & 7.67 & 4.36\\
    & mBERT & 12.34 & \textbf{6.29}$^{\dagger}$ &  10.42 & 6.70 & 4.77 & 4.69 & 3.14 & \textbf{2.55}$^{\ddag}$\\
    \midrule
    \multirow{ 2}{*}{XCOPA} & XLM-R & 20.07 & 15.95 & 5.47 & \textbf{4.52}$^{\dagger}$ & 16.02 & \textbf{4.06}$^{\ddag}$ & 8.94 & 4.39 \\
    & mBERT & 23.51 & 20.02 & 12.41 & \textbf{6.77}$^{\dagger}$ & 20.09 & 6.89 & 3.75 & \textbf{3.54}$^{\ddag}$\\
    \bottomrule

\end{tabular}
\caption{Calibration Errors ($ \mathop{\mathbb{E}}_{l \in \mathcal{L}'}[\texttt{ECE}(l)]$) for XLM-R and mBERT on using different methods for calibration. We categorize the methods into zero-shot i.e. the methods that do not use any target language data to calibrate and few-shot for the methods that require some examples in target language. Detailed results are in Table \ref{tab:det_results} of Appendix}
\label{tab:cal_impr}
\end{table*}

\subsection{Experimental Setup}
\noindent \textbf{Datasets} We consider 4 multilingual classification datasets to study calibration of MMLMs which include: i) The Cross-Lingual NLI Corpus (XNLI) \cite{Conneau2018xnli}, ii) Multilingual Dataset for Causal Commonsense Reasoning (XCOPA) \cite{ponti-etal-2020-xcopa}, iii) Multilingual Amazon Reviews Corpus (MARC) \cite{keung-et-al-2020-marc} and, iv) Cross-lingual Adversarial Dataset for Paraphrase Identification (PAWS-X) \cite{Yang2019paws-x}. Statistics of these datasets can be found in Table \ref{tab:data_stats}.
% in Appendix contains the statistics of these 4 datasets. 
% For all the datasets we consider zero-shot setting where only English data is used to fine-tune the model, and for the few-shot case we use the validation data in target languages to do continued fine-tuning. 
% We are particularly interested in XNLI and XCOPA as these two benchmarks have a much more diverse set of languages, whereas PAWS-X and MARC contains test data only in high resource languages.

\noindent\textbf{Training setup} We consider two commonly used MMLMs in our experiments i.e. Multilingual BERT (mBERT) \cite{devlin-etal-2019-bert}, and XLM-RoBERTa (XLMR) \cite{conneau-etal-2020-unsupervised}. mBERT is only available in the base variant with 12 layers and for XLMR we use the large variant with 24 layers. We use English training data to fine-tune the two MMLMs on all the tasks and evaluate the accuracies and ECEs on the test data for different languages. For the few-shot case we use the validation data in target languages to do continued fine-tuning (FSL) and temperature scaling (Self-TS). Refer to Section \ref{sec:expt_set_detail} in the Appendix for more details.

\subsection{Results}

\paragraph{Out of Box Zero-Shot Calibration (OOB)} We first investigate how well calibrated MMLMs are on the languages unseen during fine-tuning without applying any calibration techniques. As can be seen in the Table \ref{tab:oob_cali}, the average calibration error on languages other than English (column 4) is almost always significantly worse than the errors on English test data (column 3) for both mBERT and XLMR across the 4 tasks. Along with the expected calibration errors across unseen languages we also report the worst case ECE (in column 5), where we see 2$\times$ to 5$\times$ increase in errors compared to English. The worst case performance is commonly observed for low resource languages like Swahili or the languages that are typologically diverse from English like Japanese. Consequently, the overall calibration is worse on tasks like XCOPA and XNLI compared to PAWS-X and MARC, as the former two have more diverse set of languages while the latter consists of high resource languages only.

\paragraph{Factors Affecting Calibration} Next, we analyze which model-specific and typological features might influence the out-of-box calibration across languages. For a given task and MMLM, we compute Pearson correlation coefficients between the calibration errors and three factors studied extensively in zero shot cross lingual transfer literature which are:

\noindent
i) \textbf{SIZE}: Logarithm of the pre-training data size (number of tokens) of a language i.e. how well a language is represented in the pre-training corpus of an MMLM \cite{lauscher-etal-2020-zero, wu-dredze-2020-languages}, 

\noindent
ii) \textbf{SYN}: We utilize the syntactic features provided by the URIEL project \cite{littell-etal-2017-uriel} to compute the syntactic similarity between the pivot and target language as done in \citet{lin-etal-2019-choosing}.

\noindent
iii) \textbf{SWO}: Finally we consider the sub-word overlap between the pivot and target language as defined in \citet{srinivasan-et-al-2021-predicting}. To compute SWO, first vocabularies $V_p$ and $V_t$ are identified for the pivot and target langauge respectively by tokenizing the wikipedias in the two languages and getting rid of the tokens that appear less than 10 times in the corpora.  The subword overlap is then computed as :
\begin{equation*}
    \text{SWO} = \frac{|V_p \cap V_t|}{|V_p    \cup V_t|}
\end{equation*}.

In the case of XLMR, for all the tasks except MARC, we observe strong negative correlations between ECE and the three factors mentioned above (Table \ref{tab:cal_factors}), meaning lower the amount of pre-training data present for a language or its relatedness with English, the higher is the calibration error. Out of the three, the correlations are more consistently (negatively) high with SYN. We observe similar correlations albeit to a slightly lower extent for mBERT as well (Table \ref{tab:cal_factors_mbert} in Appendix).

% Finally, we also observe near perfect correlations ( > 0.95) between ECE and accuracy on the test data of different languages, meaning that calibration is worse on languages where the zero-shot performance is poor and vice versa. However, do note that this only holds true for a selected MMLM fine-tuned on a given task and not necessarily across models or task. For example, XLMR on average performs much better in terms of accuracy than mBERT on XNLI benchmark (~15\% absolute improvement in accuracy), however the ECEs obtained by both models are very close, in fact XLMR is slightly worse on average (Table \ref{tab:oob_cali}). In the following section we shall also see how we can break this correlation, i.e. reducing the calibration errors across languages without affecting the accuracy.

\paragraph{Improving Calibration} Now that we have identified miscalibration as an issue in MMLMs and factors influencing the same, we seek to improve their calibration across languages. We utilize the calibration methods described in Section \ref{sec:cal_methods}, and report the average calibration errors across the unseen languages in Table \ref{tab:cal_impr} for XNLI and XCOPA datasets \footnote{Refer to Table \ref{tab:cal_impr_mbert} in Appendix for results on MARC and PAWS-X}. Both zero-shot calibration methods (TS and LS) that only make use of English data to calibrate can be seen to obtain substantially lower calibration errors compared to out of box calibration (OOB) across all the tasks and MMLMs. Out of the two, temperature scaling often results in bigger drops in ECE compared to label smoothing with an exception of XCOPA dataset where label smoothing performs much better. In majority of the cases the errors can be reduced even further by considering the combination of the two techniques i.e. TS + LS.  
Temperature scaling by design does not affect the accuracy of uncalibrated models. In all our experiments we see that models trained with label smoothing also obtain accuracies very close to their uncalibrated counterparts. Refer to Appendix Table \ref{tab:acc} for the exact accuracy numbers.  

Next, we investigate if it is possible to reduce the calibration errors on a language even further if we are allowed to collect a few-samples in that language. We observe that when combined with label smoothing, both using  few-shot data to do temperature scaling (Self-TS + LS) or fine-tuning (FSL + LS), often results in significant drops over the errors corresponding to the best performing zero-shot calibration method. We do not use more than 2500 examples in the target language in any of the experiments, and the number of examples can be as low as 100 for XCOPA. For XNLI and XCOPA datasets, we observe that Self-TS + LS performs better than FSL+LS for XLMR and the reverse is true for mBERT. One advantage of using FSL over TS is that it can often result in increase in accuracy in addition to the reducing the calibration errors. However, we do observe an exception to this phenomenon for XCOPA where fine-tuning on the 100 validation examples hurts the overall test accuracy of the models. Hence, our general recommendation is to use FSL + LS for calibrating the models if the amount of data that can be collected is not too low, otherwise it might be more appropriate to use Self-TS + LS as learning just one parameter ($T$) should be less prone to overfitting compared to the weights of the entire network.

% \begin{table}[]
% \centering
% \small
% \begin{tabular}{p{0.9cm}p{1.1cm}p{2cm}p{2cm}}
%     \toprule
%     Dataset & MMLM & ECE ($\{en\}$) & ECE ($\mathcal{L} - \{en\}$)\\
%     \midrule
%     \multirow{ 2}{*}{XNLI} & XLM-R & 7.32 $\pm$ 1.64 & 13.34 $\pm$ 3.04\\
%     & mBERT & 5.44 $\pm$ 3.24 & 12.34 $\pm$ 10.0\\
%     \midrule
%     \multirow{ 2}{*}{XCOPA} & XLM-R & 14.54 $\pm$ 2.04 & 20.07 $\pm$ 4.78\\
%     & mBERT & 23.4 $\pm$ 13.77 & 23.51 $\pm$ 16.4\\
%     \midrule

% \end{tabular}
% \end{table}

\section{Conclusion}
In this work we showed that MMLMs like mBERT and XLMR are miscalibrated in a zero-shot cross lingual setting, with the calibration errors being even worse on low resource languages and languages that are typologically distant from the pivot language (often English). We then demonstrated the effectiveness of standard calibration techniques for improving calibration across languages both with and without collecting any new language-specific labelled data. We recommend that researchers and practitioners consider, measure and report the calibration of multilingual models while using them for scientific studies and building systems. In future work, we aim to bridge the gap between zero-shot and few-shot calibration methods by exploring unsupervised calibration methods under domain shift \cite{pampari-ermon-2020-unsupervised, pmlr-v108-park20b} that utilizes unlabelled data in new domains to improve calibration. Investigating the cross lingual calibration of MMLMs for tasks other than sentence classification like Sequence Labelling \cite{Pan2017, nivre2018universal} and Question Answering \cite{artetxe2020cross} is another natural extension of our work.

\section*{Limitations}
Our work focused on measuring and improving calibration of MMLMs across different languages and tasks. The languages that we considered in our experiments were the ones for which labelled test sets were available in the 4 multilingual benchmarks that we considered. The number of languages in these benchmarks ranged from 6 in case of MARC to 15 in XNLI, covering mostly high resource languages \footnote{class 3 or above according to the hierarchy defined by \citet{joshi-etal-2020-state}} with Swahili being the lowest resource language studied. However, the MMLMs considered in this work supports around 100 languages many of which are arguably even lower resource compared to Swahili. Hence, how well the methods discussed in the paper work towards improving the calibration for such languages needs to be explored but is  limited by the current state of multilingual benchmarking \cite{ahuja-etal-2022-beyond}.

Additionally, investigating the state of calibration across the languages for the 4 tasks and 2 MMLMs for different hyper-parameters and random seeds required a reasonably large amount of GPU resources (we used NVIDIA V100 and P100 GPUs). However, the calibration methods that we describe in the paper can work with little (for temperature scaling and few-shot learning) to no (for label smoothing) additional compute over the standard model training.

% EMNLP 2022 requires all submissions to have a section titled ``Limitations'', for discussing the limitations of the paper as a complement to the discussion of strengths in the main text. This section should occur after the conclusion, but before the references. It will not count towards the page limit.  

% The discussion of limitations is mandatory. Papers without a limitation section will be desk-rejected without review.
% ARR-reviewed papers that did not include ``Limitations'' section in their prior submission, should submit a PDF with such a section together with their EMNLP 2022 submission.

% While we are open to different types of limitations, just mentioning that a set of results have been shown for English only probably does not reflect what we expect. 
% Mentioning that the method works mostly for languages with limited morphology, like English, is a much better alternative.
% In addition, limitations such as low scalability to long text, the requirement of large GPU resources, or other things that inspire crucial further investigation are welcome.

\section*{Ethics Statement}
Our work deals with calibration of confidence predictions of classifiers trained on top of pre-trained multilingual models. Having well calibrated predictions is imperative for building robust NLP systems especially when using them for security sensitive applications like Hate Speech Detection to flag social media accounts \cite{cuthbertson-2021-AI}, decision making in law enforcement and fraud detection \cite{metz-satariano-2020-alogrithm}, where extreme actions should only be taken when we are confident about the system's prediction. However, the predicted confidences mean essentially nothing if the model is miscalibrated, posing major risks in using such models. Through our work we highlight that the commonly used multilingual models are highly miscalibrated when used in a zero-shot setting for low resource and typologically diverse languages. Additionally, we manage to substantially improve the calibration of these models across languages, addressing this linguistic disparity and boosting the reliability of such models. 
% Scientific work published at EMNLP 2022 must comply with the \href{https://www.aclweb.org/portal/content/acl-code-ethics}{ACL Ethics Policy}. We encourage all authors to include an explicit ethics statement on the broader impact of the work, or other ethical considerations after the conclusion but before the references. The ethics statement will not count toward the page limit (8 pages for long, 4 pages for short papers).

\section*{Acknowledgements}
We sincerely thank the the reviewers for their valuable feedback about our work. We would also like to thank Abhinav Kumar for brain-storming about the work and his helpful suggestions.

% Entries for the entire Anthology, followed by custom entries
\bibliography{anthology,custom}

\begin{thebibliography}{41}
\expandafter\ifx\csname natexlab\endcsname\relax\def\natexlab#1{#1}\fi

\bibitem[{Ahuja et~al.(2022)Ahuja, Dandapat, Sitaram, and
  Choudhury}]{ahuja-etal-2022-beyond}
Kabir Ahuja, Sandipan Dandapat, Sunayana Sitaram, and Monojit Choudhury. 2022.
\newblock \href {https://doi.org/10.18653/v1/2022.nlppower-1.7} {Beyond static
  models and test sets: Benchmarking the potential of pre-trained models across
  tasks and languages}.
\newblock In \emph{Proceedings of NLP Power! The First Workshop on Efficient
  Benchmarking in NLP}, pages 64--74, Dublin, Ireland. Association for
  Computational Linguistics.

\bibitem[{Ansell et~al.(2022)Ansell, Ponti, Korhonen, and
  Vuli{\'c}}]{ansell-etal-2022-composable}
Alan Ansell, Edoardo Ponti, Anna Korhonen, and Ivan Vuli{\'c}. 2022.
\newblock \href {https://doi.org/10.18653/v1/2022.acl-long.125} {Composable
  sparse fine-tuning for cross-lingual transfer}.
\newblock In \emph{Proceedings of the 60th Annual Meeting of the Association
  for Computational Linguistics (Volume 1: Long Papers)}, pages 1778--1796,
  Dublin, Ireland. Association for Computational Linguistics.

\bibitem[{Artetxe et~al.(2020)Artetxe, Ruder, and Yogatama}]{artetxe2020cross}
Mikel Artetxe, Sebastian Ruder, and Dani Yogatama. 2020.
\newblock {On the Cross-lingual Transferability of Monolingual
  Representations}.
\newblock In \emph{Proceedings of ACL 2020}.

\bibitem[{Chen et~al.(2017)Chen, Zhu, Ling, Wei, Jiang, and
  Inkpen}]{chen-etal-2017-enhanced}
Qian Chen, Xiaodan Zhu, Zhen-Hua Ling, Si~Wei, Hui Jiang, and Diana Inkpen.
  2017.
\newblock \href {https://doi.org/10.18653/v1/P17-1152} {Enhanced {LSTM} for
  natural language inference}.
\newblock In \emph{Proceedings of the 55th Annual Meeting of the Association
  for Computational Linguistics (Volume 1: Long Papers)}, pages 1657--1668,
  Vancouver, Canada. Association for Computational Linguistics.

\bibitem[{Conneau et~al.(2020)Conneau, Khandelwal, Goyal, Chaudhary, Wenzek,
  Guzm{\'a}n, Grave, Ott, Zettlemoyer, and
  Stoyanov}]{conneau-etal-2020-unsupervised}
Alexis Conneau, Kartikay Khandelwal, Naman Goyal, Vishrav Chaudhary, Guillaume
  Wenzek, Francisco Guzm{\'a}n, Edouard Grave, Myle Ott, Luke Zettlemoyer, and
  Veselin Stoyanov. 2020.
\newblock \href {https://doi.org/10.18653/v1/2020.acl-main.747} {Unsupervised
  cross-lingual representation learning at scale}.
\newblock In \emph{Proceedings of the 58th Annual Meeting of the Association
  for Computational Linguistics}, pages 8440--8451, Online. Association for
  Computational Linguistics.

\bibitem[{Conneau et~al.(2018)Conneau, Rinott, Lample, Williams, Bowman,
  Schwenk, and Stoyanov}]{Conneau2018xnli}
Alexis Conneau, Ruty Rinott, Guillaume Lample, Adina Williams, Samuel Bowman,
  Holger Schwenk, and Veselin Stoyanov. 2018.
\newblock {XNLI}: Evaluating cross-lingual sentence representations.
\newblock In \emph{Proceedings of EMNLP 2018}, pages 2475--2485.

\bibitem[{Cuthbertson(2021)}]{cuthbertson-2021-AI}
Anthony Cuthbertson. 2021.
\newblock \href
  {https://www.independent.co.uk/tech/ai-chess-racism-youtube-agadmator-b1804160.html}
  {Ai mistakes ‘black and white’ chess chat for racism}.
\newblock \emph{The Independent}.

\bibitem[{Davidson et~al.(2017)Davidson, Warmsley, Macy, and
  Weber}]{DavidsonWMW17}
Thomas Davidson, Dana Warmsley, Michael~W. Macy, and Ingmar Weber. 2017.
\newblock \href {https://aaai.org/ocs/index.php/ICWSM/ICWSM17/paper/view/15665}
  {Automated hate speech detection and the problem of offensive language}.
\newblock In \emph{ICWSM}, pages 512--515.

\bibitem[{Desai and Durrett(2020)}]{desai-durrett-2020-calibration}
Shrey Desai and Greg Durrett. 2020.
\newblock \href {https://doi.org/10.18653/v1/2020.emnlp-main.21} {Calibration
  of pre-trained transformers}.
\newblock In \emph{Proceedings of the 2020 Conference on Empirical Methods in
  Natural Language Processing (EMNLP)}, pages 295--302, Online. Association for
  Computational Linguistics.

\bibitem[{Deshpande et~al.(2022)Deshpande, Farris, and
  Kumar}]{deshpande-2022-highly}
Neha Deshpande, Nicholas Farris, and Vidhur Kumar. 2022.
\newblock \href {https://doi.org/10.48550/ARXIV.2201.11294} {Highly
  generalizable models for multilingual hate speech detection}.

\bibitem[{Devlin et~al.(2019)Devlin, Chang, Lee, and
  Toutanova}]{devlin-etal-2019-bert}
Jacob Devlin, Ming-Wei Chang, Kenton Lee, and Kristina Toutanova. 2019.
\newblock \href {https://doi.org/10.18653/v1/N19-1423} {{BERT}: Pre-training of
  deep bidirectional transformers for language understanding}.
\newblock In \emph{Proceedings of the 2019 Conference of the North {A}merican
  Chapter of the Association for Computational Linguistics: Human Language
  Technologies, Volume 1 (Long and Short Papers)}, pages 4171--4186,
  Minneapolis, Minnesota. Association for Computational Linguistics.

\bibitem[{Guo et~al.(2017)Guo, Pleiss, Sun, and Weinberger}]{guo-2017-on}
Chuan Guo, Geoff Pleiss, Yu~Sun, and Kilian~Q. Weinberger. 2017.
\newblock \href {https://proceedings.mlr.press/v70/guo17a.html} {On calibration
  of modern neural networks}.
\newblock In \emph{Proceedings of the 34th International Conference on Machine
  Learning}, volume~70 of \emph{Proceedings of Machine Learning Research},
  pages 1321--1330. PMLR.

\bibitem[{He et~al.(2016)He, Zhang, Ren, and Sun}]{he-et-al-2016-deep}
Kaiming He, Xiangyu Zhang, Shaoqing Ren, and Jian Sun. 2016.
\newblock \href {https://doi.org/10.1109/CVPR.2016.90} {Deep residual learning
  for image recognition}.
\newblock In \emph{2016 IEEE Conference on Computer Vision and Pattern
  Recognition (CVPR)}, pages 770--778.

\bibitem[{Joshi et~al.(2020)Joshi, Santy, Budhiraja, Bali, and
  Choudhury}]{joshi-etal-2020-state}
Pratik Joshi, Sebastin Santy, Amar Budhiraja, Kalika Bali, and Monojit
  Choudhury. 2020.
\newblock \href {https://doi.org/10.18653/v1/2020.acl-main.560} {The state and
  fate of linguistic diversity and inclusion in the {NLP} world}.
\newblock In \emph{Proceedings of the 58th Annual Meeting of the Association
  for Computational Linguistics}, pages 6282--6293, Online. Association for
  Computational Linguistics.

\bibitem[{Keung et~al.(2020)Keung, Lu, Szarvas, and
  Smith}]{keung-et-al-2020-marc}
Phillip Keung, Yichao Lu, Gy{\"{o}}rgy Szarvas, and Noah~A. Smith. 2020.
\newblock \href {http://arxiv.org/abs/2010.02573} {The multilingual amazon
  reviews corpus}.
\newblock \emph{CoRR}, abs/2010.02573.

\bibitem[{Kingma and Ba(2015)}]{kingma-ba-2015-adam}
Diederik~P. Kingma and Jimmy Ba. 2015.
\newblock \href {http://arxiv.org/abs/1412.6980} {Adam: A method for stochastic
  optimization}.
\newblock In \emph{ICLR (Poster)}.

\bibitem[{Lauscher et~al.(2020)Lauscher, Ravishankar, Vuli{\'c}, and
  Glava{\v{s}}}]{lauscher-etal-2020-zero}
Anne Lauscher, Vinit Ravishankar, Ivan Vuli{\'c}, and Goran Glava{\v{s}}. 2020.
\newblock \href {https://doi.org/10.18653/v1/2020.emnlp-main.363} {From zero to
  hero: {O}n the limitations of zero-shot language transfer with multilingual
  {T}ransformers}.
\newblock In \emph{Proceedings of the 2020 Conference on Empirical Methods in
  Natural Language Processing (EMNLP)}, pages 4483--4499, Online. Association
  for Computational Linguistics.

\bibitem[{Lecun et~al.(1998)Lecun, Bottou, Bengio, and
  Haffner}]{lecun-et-al-1998-gradient}
Y.~Lecun, L.~Bottou, Y.~Bengio, and P.~Haffner. 1998.
\newblock \href {https://doi.org/10.1109/5.726791} {Gradient-based learning
  applied to document recognition}.
\newblock \emph{Proceedings of the IEEE}, 86(11):2278--2324.

\bibitem[{Lin et~al.(2019)Lin, Chen, Lee, Li, Zhang, Xia, Rijhwani, He, Zhang,
  Ma, Anastasopoulos, Littell, and Neubig}]{lin-etal-2019-choosing}
Yu-Hsiang Lin, Chian-Yu Chen, Jean Lee, Zirui Li, Yuyan Zhang, Mengzhou Xia,
  Shruti Rijhwani, Junxian He, Zhisong Zhang, Xuezhe Ma, Antonios
  Anastasopoulos, Patrick Littell, and Graham Neubig. 2019.
\newblock \href {https://doi.org/10.18653/v1/P19-1301} {Choosing transfer
  languages for cross-lingual learning}.
\newblock In \emph{Proceedings of the 57th Annual Meeting of the Association
  for Computational Linguistics}, pages 3125--3135, Florence, Italy.
  Association for Computational Linguistics.

\bibitem[{Littell et~al.(2017)Littell, Mortensen, Lin, Kairis, Turner, and
  Levin}]{littell-etal-2017-uriel}
Patrick Littell, David~R. Mortensen, Ke~Lin, Katherine Kairis, Carlisle Turner,
  and Lori Levin. 2017.
\newblock \href {https://aclanthology.org/E17-2002} {{URIEL} and lang2vec:
  Representing languages as typological, geographical, and phylogenetic
  vectors}.
\newblock In \emph{Proceedings of the 15th Conference of the {E}uropean Chapter
  of the Association for Computational Linguistics: Volume 2, Short Papers},
  pages 8--14, Valencia, Spain. Association for Computational Linguistics.

\bibitem[{Liu et~al.(2020)Liu, Gu, Goyal, Li, Edunov, Ghazvininejad, Lewis, and
  Zettlemoyer}]{liu-etal-2020-multilingual-denoising}
Yinhan Liu, Jiatao Gu, Naman Goyal, Xian Li, Sergey Edunov, Marjan
  Ghazvininejad, Mike Lewis, and Luke Zettlemoyer. 2020.
\newblock \href {https://doi.org/10.1162/tacl_a_00343} {Multilingual denoising
  pre-training for neural machine translation}.
\newblock \emph{Transactions of the Association for Computational Linguistics},
  8:726--742.

\bibitem[{Metz and Satariano(2020)}]{metz-satariano-2020-alogrithm}
Cade Metz and Adam Satariano. 2020.
\newblock \href
  {https://www.nytimes.com/2020/02/06/technology/predictive-algorithms-crime.html}
  {An algorithm that grants freedom, or takes it away.}
\newblock \emph{The New York Times}.

\bibitem[{M\"{u}ller et~al.(2019)M\"{u}ller, Kornblith, and
  Hinton}]{rafael-et-al-2019-when}
Rafael M\"{u}ller, Simon Kornblith, and Geoffrey~E Hinton. 2019.
\newblock \href
  {https://proceedings.neurips.cc/paper/2019/file/f1748d6b0fd9d439f71450117eba2725-Paper.pdf}
  {When does label smoothing help?}
\newblock In \emph{Advances in Neural Information Processing Systems},
  volume~32. Curran Associates, Inc.

\bibitem[{Nivre et~al.(2018)Nivre, Abrams, Agi{\'c}, Ahrenberg, Antonsen,
  Aranzabe, Arutie, Asahara, Ateyah, Attia et~al.}]{nivre2018universal}
Joakim Nivre, Mitchell Abrams, {\v{Z}}eljko Agi{\'c}, Lars Ahrenberg, Lene
  Antonsen, Maria~Jesus Aranzabe, Gashaw Arutie, Masayuki Asahara, Luma Ateyah,
  Mohammed Attia, et~al. 2018.
\newblock Universal dependencies 2.2.

\bibitem[{Pampari and Ermon(2020)}]{pampari-ermon-2020-unsupervised}
Anusri Pampari and Stefano Ermon. 2020.
\newblock \href {http://arxiv.org/abs/2006.16405} {Unsupervised calibration
  under covariate shift}.
\newblock \emph{CoRR}, abs/2006.16405.

\bibitem[{Pan et~al.(2017)Pan, Zhang, May, Nothman, Knight, and Ji}]{Pan2017}
Xiaoman Pan, Boliang Zhang, Jonathan May, Joel Nothman, Kevin Knight, and Heng
  Ji. 2017.
\newblock {Cross-lingual name tagging and linking for 282 languages}.
\newblock In \emph{Proceedings of ACL 2017}, pages 1946--1958.

\bibitem[{Park et~al.(2020)Park, Bastani, Weimer, and Lee}]{pmlr-v108-park20b}
Sangdon Park, Osbert Bastani, James Weimer, and Insup Lee. 2020.
\newblock \href {https://proceedings.mlr.press/v108/park20b.html} {Calibrated
  prediction with covariate shift via unsupervised domain adaptation}.
\newblock In \emph{Proceedings of the Twenty Third International Conference on
  Artificial Intelligence and Statistics}, volume 108 of \emph{Proceedings of
  Machine Learning Research}, pages 3219--3229. PMLR.

\bibitem[{Pereyra et~al.(2017)Pereyra, Tucker, Chorowski, Kaiser, and
  Hinton}]{pereyra-et-al-2017}
Gabriel Pereyra, George Tucker, Jan Chorowski, Łukasz Kaiser, and Geoffrey
  Hinton. 2017.
\newblock \href {https://doi.org/10.48550/ARXIV.1701.06548} {Regularizing
  neural networks by penalizing confident output distributions}.

\bibitem[{Pfeiffer et~al.(2020)Pfeiffer, Vuli{\'c}, Gurevych, and
  Ruder}]{pfeiffer-etal-2020-mad}
Jonas Pfeiffer, Ivan Vuli{\'c}, Iryna Gurevych, and Sebastian Ruder. 2020.
\newblock \href {https://doi.org/10.18653/v1/2020.emnlp-main.617} {{MAD-X}:
  {A}n {A}dapter-{B}ased {F}ramework for {M}ulti-{T}ask {C}ross-{L}ingual
  {T}ransfer}.
\newblock In \emph{Proceedings of the 2020 Conference on Empirical Methods in
  Natural Language Processing (EMNLP)}, pages 7654--7673, Online. Association
  for Computational Linguistics.

\bibitem[{Ponti et~al.(2021)Ponti, Vuli{\'c}, Cotterell, Parovic, Reichart, and
  Korhonen}]{ponti-etal-2021-parameter}
Edoardo~M. Ponti, Ivan Vuli{\'c}, Ryan Cotterell, Marinela Parovic, Roi
  Reichart, and Anna Korhonen. 2021.
\newblock \href {https://doi.org/10.1162/tacl_a_00374} {Parameter space
  factorization for zero-shot learning across tasks and languages}.
\newblock \emph{Transactions of the Association for Computational Linguistics},
  9:410--428.

\bibitem[{Ponti et~al.(2020)Ponti, Glava{\v{s}}, Majewska, Liu, Vuli{\'c}, and
  Korhonen}]{ponti-etal-2020-xcopa}
Edoardo~Maria Ponti, Goran Glava{\v{s}}, Olga Majewska, Qianchu Liu, Ivan
  Vuli{\'c}, and Anna Korhonen. 2020.
\newblock \href {https://doi.org/10.18653/v1/2020.emnlp-main.185} {{XCOPA}: A
  multilingual dataset for causal commonsense reasoning}.
\newblock In \emph{Proceedings of the 2020 Conference on Empirical Methods in
  Natural Language Processing (EMNLP)}, pages 2362--2376, Online. Association
  for Computational Linguistics.

\bibitem[{Roemmele et~al.(2011)Roemmele, Bejan, and
  Gordon}]{roemmele2011choice}
Melissa Roemmele, Cosmin~Adrian Bejan, and Andrew~S Gordon. 2011.
\newblock Choice of plausible alternatives: An evaluation of commonsense causal
  reasoning.
\newblock In \emph{AAAI spring symposium: logical formalizations of commonsense
  reasoning}, pages 90--95.

\bibitem[{Sap et~al.(2019)Sap, Rashkin, Chen, Le~Bras, and
  Choi}]{sap-etal-2019-social}
Maarten Sap, Hannah Rashkin, Derek Chen, Ronan Le~Bras, and Yejin Choi. 2019.
\newblock \href {https://doi.org/10.18653/v1/D19-1454} {Social {IQ}a:
  Commonsense reasoning about social interactions}.
\newblock In \emph{Proceedings of the 2019 Conference on Empirical Methods in
  Natural Language Processing and the 9th International Joint Conference on
  Natural Language Processing (EMNLP-IJCNLP)}, pages 4463--4473, Hong Kong,
  China. Association for Computational Linguistics.

\bibitem[{Sarkar and KhudaBukhsh(2021)}]{Sarkar_KhudaBukhsh_2021}
Rupak Sarkar and Ashiqur~R. KhudaBukhsh. 2021.
\newblock \href {https://ojs.aaai.org/index.php/AAAI/article/view/17937} {Are
  chess discussions racist? an adversarial hate speech data set (student
  abstract)}.
\newblock \emph{Proceedings of the AAAI Conference on Artificial Intelligence},
  35(18):15881--15882.

\bibitem[{Srinivasan et~al.(2021)Srinivasan, Sitaram, Ganu, Dandapat, Bali, and
  Choudhury}]{srinivasan-et-al-2021-predicting}
Anirudh Srinivasan, Sunayana Sitaram, Tanuja Ganu, Sandipan Dandapat, Kalika
  Bali, and Monojit Choudhury. 2021.
\newblock \href {https://doi.org/10.48550/ARXIV.2110.08875} {Predicting the
  performance of multilingual nlp models}.

\bibitem[{Williams et~al.(2018)Williams, Nangia, and
  Bowman}]{williams-etal-2018-broad}
Adina Williams, Nikita Nangia, and Samuel Bowman. 2018.
\newblock \href {https://doi.org/10.18653/v1/N18-1101} {A broad-coverage
  challenge corpus for sentence understanding through inference}.
\newblock In \emph{Proceedings of the 2018 Conference of the North {A}merican
  Chapter of the Association for Computational Linguistics: Human Language
  Technologies, Volume 1 (Long Papers)}, pages 1112--1122, New Orleans,
  Louisiana. Association for Computational Linguistics.

\bibitem[{Wolf et~al.(2020)Wolf, Debut, Sanh, Chaumond, Delangue, Moi, Cistac,
  Rault, Louf, Funtowicz, Davison, Shleifer, von Platen, Ma, Jernite, Plu, Xu,
  Le~Scao, Gugger, Drame, Lhoest, and Rush}]{wolf-etal-2020-transformers}
Thomas Wolf, Lysandre Debut, Victor Sanh, Julien Chaumond, Clement Delangue,
  Anthony Moi, Pierric Cistac, Tim Rault, Remi Louf, Morgan Funtowicz, Joe
  Davison, Sam Shleifer, Patrick von Platen, Clara Ma, Yacine Jernite, Julien
  Plu, Canwen Xu, Teven Le~Scao, Sylvain Gugger, Mariama Drame, Quentin Lhoest,
  and Alexander Rush. 2020.
\newblock \href {https://doi.org/10.18653/v1/2020.emnlp-demos.6} {Transformers:
  State-of-the-art natural language processing}.
\newblock In \emph{Proceedings of the 2020 Conference on Empirical Methods in
  Natural Language Processing: System Demonstrations}, pages 38--45, Online.
  Association for Computational Linguistics.

\bibitem[{Wu and Dredze(2020)}]{wu-dredze-2020-languages}
Shijie Wu and Mark Dredze. 2020.
\newblock \href {https://doi.org/10.18653/v1/2020.repl4nlp-1.16} {Are all
  languages created equal in multilingual {BERT}?}
\newblock In \emph{Proceedings of the 5th Workshop on Representation Learning
  for NLP}, pages 120--130, Online. Association for Computational Linguistics.

\bibitem[{Xue et~al.(2021)Xue, Constant, Roberts, Kale, Al-Rfou, Siddhant,
  Barua, and Raffel}]{xue-etal-2021-mt5}
Linting Xue, Noah Constant, Adam Roberts, Mihir Kale, Rami Al-Rfou, Aditya
  Siddhant, Aditya Barua, and Colin Raffel. 2021.
\newblock \href {https://doi.org/10.18653/v1/2021.naacl-main.41} {m{T}5: A
  massively multilingual pre-trained text-to-text transformer}.
\newblock In \emph{Proceedings of the 2021 Conference of the North American
  Chapter of the Association for Computational Linguistics: Human Language
  Technologies}, pages 483--498, Online. Association for Computational
  Linguistics.

\bibitem[{Yang et~al.(2019)Yang, Zhang, Tar, and Baldridge}]{Yang2019paws-x}
Yinfei Yang, Yuan Zhang, Chris Tar, and Jason Baldridge. 2019.
\newblock {PAWS-X}: A cross-lingual adversarial dataset for paraphrase
  identification.
\newblock In \emph{Proceedings of EMNLP 2019}, pages 3685--3690.

\bibitem[{Zhang et~al.(2019)Zhang, Baldridge, and He}]{zhang-etal-2019-paws}
Yuan Zhang, Jason Baldridge, and Luheng He. 2019.
\newblock \href {https://doi.org/10.18653/v1/N19-1131} {{PAWS}: Paraphrase
  adversaries from word scrambling}.
\newblock In \emph{Proceedings of the 2019 Conference of the North {A}merican
  Chapter of the Association for Computational Linguistics: Human Language
  Technologies, Volume 1 (Long and Short Papers)}, pages 1298--1308,
  Minneapolis, Minnesota. Association for Computational Linguistics.

\end{thebibliography}
\bibliographystyle{acl_natbib}

\appendix

\clearpage
\section{Appendix}
\label{sec:appendix}

\subsection{Measuring Calibration}
\label{sec:cal_measure}
\paragraph{Reliability Diagrams} are an effective way of visualizing the calibration of a classifier. To plot these, we first predict the (maximum) confidence for each example in the test set and group the data points into $M$ equally sized bins based on the predicted confidence. For each bin, accuracy is computed (by considering label with maximum confidence as prediction) and plotted on the y axis with confidence being on the x-axis, as can be seen in Figure \ref{fig:cal_en_sw} (blue bars). For a perfectly calibrated classifier the confidence of a bin should match with the classifier's accuracy on that bin i.e. the accuracy vs confidence curve should lie on the line $y = x$ (red-dotted line in Figure \ref{fig:cal_en_sw}). The gap between the two is plotted (as red bars in Figure \ref{fig:cal_en_sw}) to represent the calibration error of the classifier. 

\paragraph{Expected Calibration Error (ECE)} is defined as expected value of the difference between the confidence and accuracy of the classifier's predictions. In practice ECE of a classifier is computed using a binning strategy as described in \citet{guo-2017-on} 
where the confidence predictions (corresponding to the maximum confidence class) on the $n$ test examples are grouped into $M$ uniform-sized bins, such that set of examples belonging to the $m^{th}$ bin are denoted by $B_m$. Accuracy ($\text{acc}(B_m)$) and average confidence ($\text{conf}(B_m)$) for each bin is computed and a weighted average of the differences between the two is taken to obtain ECE.
\begin{equation*}
    \texttt{ECE} =  \sum_{m = 1}^M \frac{|B_m|}{n} |\text{acc}(B_m) - \text{conf}(B_m)|
\end{equation*}

\subsection{Datasets Description}

\noindent\textbf{1. XNLI}\footnote{\url{https://github.com/facebookresearch/XNLI}} \cite{Conneau2018xnli}  is a Natural Langauge Inference task where a premise and hypothesis are given and the task is to predict if the hypothesis is \textit{entailed} in premise,  \textit{contradicts} the premise, or is \textit{neutral} towards it, hence being a three way classification problem. MultiNLI \cite{williams-etal-2018-broad} corpus which is available in English is used as training set, and dev and validation sets are obtained by manually translating crowd sourced English sentences for the task into 14 other languages (Arabic, Bulgarian, German, Greek, Spanish, French, Hindi, Russian, Swahili, Thai, Turkish, Urdu, Vietnamese and Chinese.)

\noindent\textbf{2. XCOPA}\footnote{\url{https://huggingface.co/datasets/xcopa}} \cite{ponti-etal-2020-xcopa} is a multilingual benchmark for causal commonsense reasoning. It was obtained by extending the dev and test sets of Choice of Plausible Alternatives (COPA)\footnote{\url{https://huggingface.co/datasets/super_glue}} \cite{roemmele2011choice} dataset to  11 typologically diverse languages. The task here is, given a premise and two alternatives, predict which alternative has a causal relationship with the premise. The original COPA dataset has only 400 training examples in English, hence it is common to first train the model on Social-IQA (SIQA)\footnote{\url{https://huggingface.co/datasets/social_i_qa}} \cite{sap-etal-2019-social} dataset (which has around 33k training examples), and then fine-tune it on COPA, and that's the strategy we adopt in our experiments. SIQA is similar to COPA but the questions or premise are defined in such a way that the possible answers have social implications, and instead of two alternatives it has three. Out of the 11 supported languages we experiment with 9 languages, ignoring Quechua and Haitian Creole as both mBERT and XLM-R were not pre-trained on these languages, leaving us with Greek, Indonesian, Italian, Swahili, Tamil, Thai, Turkish, Vietnamese and Chinese.

\noindent\textbf{3. MARC}\footnote{\url{https://registry.opendata.aws/amazon-reviews-ml/}} \cite{keung-et-al-2020-marc} is the multilingual Amazon product reviews corpus. We are given title, body and category of the review and the task is to predict the corresponding rating from 1 to 5. The corpus contains train, dev and test sets in six high resource languages: English, German, Spanish, French, Japanese and Chinese. In our experiments we only fine-tune using title and body text and ignore the category information.

\noindent\textbf{4. PAWS-X}\footnote{\url{https://huggingface.co/datasets/paws-x}} \cite{Yang2019paws-x} is an adversarial dataset for multilingual paraphrase detection. It was adapted from PAWS dataset \cite{zhang-etal-2019-paws} by manually translating the dev and test sets in English to six high resource languages (French, German, Spanish, Chinese, Japanese and Korean). The task is, given a pair of sentences predict whether the two are paraphrases of each other. For training the original English PAWS dataset is used. 

The statistics of all these datasets are provided in Table \ref{tab:data_stats}

\subsection{Detailed Experimental Setup}
\label{sec:expt_set_detail}
For fine-tuning the model we do a grid search over the learning rate($[1\text{e-}5, 3\text{e-}5, 5\text{e-}6, 7\text{e-}6]$) and the number of epochs ($[1, 3, 4, 5]$), run each setting for 3 random seeds (1, 11 and 22) and select the best hyper-parameter set corresponding to the average dev accuracy on English data \footnote{For XCOPA we did use dev data in all languages for model selection as we saw performance in English data to not necessarily correlate well with performance on other languages.}. 
% For evaluating on XCOPA dataset we first fine-tune the MMLMs on SocialIQA (SIQA) \cite{sap-etal-2019-social} dataset and then do continued fine-tuning on Choice of Plausible Alternatives (COPA) \cite{roemmele2011choice} dataset.
The final set of hyperparameters used for each dataset and MMLM are provided in Table \ref{tab:hyps}. Apart from these we use a batch size of 8 in all experiments. We use Adam optimizer \cite{kingma-ba-2015-adam} to train all of our models and LBFGS to learn the temperature parameter while performing temperature scaling. 

For computing ECE values for different tasks and languages we use $M = 10$ i.e. 10 buckets. For label smoothing we set the smoothing parameter $\alpha = 0.1$ in all experiments and initialize the temperature $T = 1.5$ while doing temperature scaling. For few-shot cases we use $\min(2500, |\mathcal{D}_{val}(\mathfrak{t}, l)|)$ examples, where $ |\mathcal{D}_{val}(\mathfrak{t}, l)|$ denotes the number of dev examples available in task $\mathfrak{t}$ for language $l$, and use that to perform continued fine-tuning or temperature scaling.

All the experiments were run on NVIDIA V100 and P100 GPUs with 32GB and 16GB memory respectively. We use pre-trained models available in Hugging Face's Transformers library \cite{wolf-etal-2020-transformers}. For computing the calibration errors as well as plotting the reliability diagrams we use the open source tool Calibration Framework\footnote{\url{https://github.com/fabiankueppers/calibration-framework}}. To encourage the research in this area we will make our code public.

\begin{table*}[]
    \small
    \centering
    \begin{tabular}{p{1cm}p{1.5cm}p{2.5cm}p{2cm}p{1.4cm}p{1.4cm}}
         \toprule
         Dataset & MMLM & Learning Rate & Epochs & Few-Shot Learning Rate & Few-Shot Epochs  \\
         \midrule
         \multirow{ 2}{*}{XNLI} & XLM-R & 7e-6 & 3 & 5e-06 & 2 \\
         & mBERT & 3e-5 & 3 & 3e-5 & 1\\
         \midrule
         \multirow{ 2}{*}{XCOPA} & XLM-R & 5e-6 (for SIQA) \& 5e-6 (for COPA)  & 4 (for S-IQA) \& 10 (for COPA) & 1e-5 & 1 \\
         & mBERT & 1e-5 (for S-IQA) \& 3e-5 (for COPA) & 3 (for SIQA) \& 10 (for COPA) & 3e-5 & 10\\
         \midrule
         \multirow{ 2}{*}{MARC} & XLM-R & 5e-6 & 3 & 5e-6 & 1 \\
         & mBERT & 5e-6 & 3 & 5e-6 & 1\\
         \midrule
         \multirow{ 2}{*}{PAWS-X} & XLM-R & 7e-6 & 3 & 3e-5 & 1 \\
         & mBERT & 1e-5 & 3 & 3e-5 & 1\\
         \bottomrule
    \end{tabular}
    \caption{Final list of hyperparameters used for reporting results.}
    \label{tab:hyps}
\end{table*}

% For calibrating the models with label smoothing, we chose $\alpha = 0.1$ and accordingly modify the fine-tuning objective. On the other hand, methods like temperature scaling and few-shot learning are applied post-hoc after fine-tuning on English is done and dev data in English (for TS) or the target languages is used to perform calibration (for Self-TS and FSL).  We also consider combinations of different calibration methods in our experiments, including Temperature Scaling + Label Smoothing (TS + LS or Self-TS + LS) and Few-Shot Learning + Label Smoothing (FSL + LS) in addition to using these methods in isolation.
% \footnote{We experimented with Few-Shot Learning + Temperature Scaling as well i.e. FSL + TS but found this combination to be highly unstable.} in addition to using these methods in isolation.

% Details about the final set of hyper-parameters for each task, MMLM and calibration method are given in Appendix \myworries{Add hyperparameter table in appendix and put a reference here}.

% \paragraph{Calibration Settings}  We also consider combinations of different calibration methods in our experiments, including Temperature Scaling + Label Smoothing (TS + LS or Self-TS + LS) and Few-Shot Learning + Label Smoothing (FSL + LS) \footnote{We experimented with Few-Shot Learning + Temperature Scaling as well i.e. FSL + TS but found this combination to be highly unstable.} in addition to using these methods in isolation.

\begin{table*}[]
    \small
    \centering
    \begin{tabular}{p{3cm}p{1.5cm}p{1.2cm}p{2cm}p{1.4cm}p{1.4cm}}
         \toprule
         Dataset & Number of Languages & Number of Labels & Training Size & Dev Size & Test Size  \\
         \midrule
         XNLI (sub-sampled) & 15 & 3 & 40000 & 2500 & 2500\\
         XCOPA & 9 & 2 & 33410 (S-IQA) + 500 (COPA) & 100 & 400\\
         MARC (sub-sampled) & 6 & 5 & 40000 & 2500 & 5000\\
         PAWS-X & 7 & 2 & 50000 & 2000 & 2000\\
         \bottomrule
    \end{tabular}
    \caption{Dataset statistics for the 4 multilingual classification tasks that we study in our experiments. We use sub-sampled versions of XNLI and MARC and only use the first 40k examples in their training sets to reduce the compute overhead and making the scale of training data is consistent across all the tasks. For completeness we also run some preliminary experiments with using the entire data for fine-tuning XNLI and present the calibration errors in Figure \ref{fig:full_v_sample}.}
    \label{tab:data_stats}
\end{table*}

\begin{table}[]
    \centering
    \small
    \begin{tabular}{p{2.5cm}p{1cm}p{1cm}p{1cm}}
        \toprule
         Dataset & SIZE & SYN & SWO \\
         \midrule
         XNLI & -0.4 & -0.38 & \textbf{-0.41} \\
         XNLI (wo th) & -0.86 & \textbf{-0.88} & -0.78 \\
         XCOPA & -0.14 & \textbf{-0.22} & 0.07 \\
         MARC & -0.5 & -0.69 & \textbf{-0.86} \\
         PAWS-X & -0.91 & -0.91 & \textbf{-0.98} \\
         \bottomrule
    \end{tabular}
    \caption{Pearson correlation coefficient between the
Expected Calibration Error (ECE) and SIZE, SYN, and
SWO features of different languages in the test set for
mBERT}
    \label{tab:cal_factors_mbert}
\end{table}

\begin{table*}[]
\small
\centering
\begin{tabular}{p{1cm}p{1.2cm}p{0.9cm}p{0.9cm}p{0.9cm}p{1.2cm}p{1.2cm}p{1.7cm}p{0.9cm}p{1.2cm}}
    \toprule
    \multirow{ 2}{*}{Dataset} & \multirow{ 2}{*}{MMLM} & \multicolumn{4}{c}{Zero-Shot Calibration} & \multicolumn{4}{c}{Few-Shot Calibration}\\
    \cmidrule(l){3-6}  \cmidrule(l){7-10}
    & & OOB & TS & LS & TS + LS & Self-TS & Self-TS + LS & FSL & FSL + LS\\
    \cmidrule(l){3-10}
    \multirow{ 2}{*}{MARC} & XLM-R & 9.65 & \textbf{4.22}$^{\dagger}$ & 7.93 & 4.45 & 3.55 & 3.36 & \textbf{2.51}$^{\ddag}$ & 2.58 \\
    & mBERT & 11.11 & 6.14 & 5.96 & \textbf{4.46}$^{\dagger}$ & 4.62 & 4.71 & \textbf{2.12}$^{\ddag}$ & 3.01\\
    \midrule
    \multirow{ 2}{*}{PAWS-X} & XLM-R & 4.28 & \textbf{2.29}$^{\dagger}$ & 3.37 & 4.44 & - & - & - & - \\
    & mBERT & 10.33 & 5.64 & 5.58 & \textbf{5.36}$^{\dagger}$ & - & - & - & -\\
    \bottomrule

\end{tabular}
\caption{Calibration Errors across tasks for XLM-R and mBERT on using different methods for calibration. For PAWS-X we only report numbers for Zero-Shot methods as the dev data and test data of the benchmark have sentences in common (even though the pairs are unique), hence we avoid using dev examples as few-shot in this case.}
\label{tab:cal_impr_mbert}
\end{table*}

\begin{table*}[]
\small
\centering
\begin{tabular}{p{1cm}p{1.2cm}p{0.9cm}p{0.9cm}p{0.9cm}p{1.2cm}p{1.2cm}p{1.7cm}p{0.9cm}p{1.2cm}}
    \toprule
    \multirow{ 2}{*}{Dataset} & \multirow{ 2}{*}{MMLM} & \multicolumn{4}{c}{Zero-Shot Calibration} & \multicolumn{4}{c}{Few-Shot Calibration}\\
    \cmidrule(l){3-6}  \cmidrule(l){7-10}
    & & OOB & TS & LS & TS + LS & Self-TS & Self-TS + LS & FSL & FSL + LS\\
    \cmidrule(l){3-10}
    \multirow{ 2}{*}{XNLI} & XLM-R & 74.9 & 74.9 &  74.6 & 74.6 & 74.9 & 74.6 & 79.4 & 79.3\\
    & mBERT & 56.7 & 56.7 &  57.5 & 57.5 & 56.7 & 57.5 & 60.6 & {61.2}\\
    \midrule
    \multirow{ 2}{*}{XCOPA} & XLM-R & 74.3 & 74.3 & {75.1} & {75.1} & 74.3 & {75.1} & 67.2 & 69.5 \\
    & mBERT & 54.5 & 54.5 & 54.4 & 54.4 & 54.5 & 54.4 & 53.0 & 52.9\\
    \midrule
    \multirow{ 2}{*}{MARC} & XLM-R & 57.7 & 57.7 & 57.6 & 57.6 & 57.7 & 57.6 & {59.4} & 59 \\
    & mBERT & 42.9 & 42.9 & 42.6 & 42.6 & 42.9 & 42.6 & {49.6} & 49.2\\
    \midrule
    \multirow{ 2}{*}{PAWS-X} & XLM-R & 76.1 & 76.1 & 75.6 & 75.6 & - & - & - & - \\
    & mBERT & 80.4 & 80.4 & 81.1 & 81.1 & - & - & - & -\\
    \bottomrule

\end{tabular}
\vspace*{-3mm}
\caption{Accuracy ($ \mathop{\mathbb{E}}_{l \in \mathcal{L}'}[\texttt{Accuracy}(l)]$) for XLM-R and mBERT on using different methods for calibration. Similar to Table \ref{tab:cal_impr_mbert}, here again we skip few-shot calibration for PAWS-X due to the possible data leakage.}
\vspace*{-3mm}
\label{tab:acc}
\end{table*}

\begin{table*}[h]
    \centering
    \small
    \begin{subtable}[h]{0.45\textwidth}
        \begin{tabular}{p{1.7cm}p{1cm}p{1.5cm}p{1.5cm}}
            \toprule
             Method & $\texttt{ECE}(en)$ & $\displaystyle \mathop{\mathbb{E}}_{l \in \mathcal{L}'}[\texttt{ECE}(l)]$ & $\displaystyle \max_{l \in \mathcal{L}'}\texttt{ECE}(l)$\\
            \midrule
            \multicolumn{4}{c}{Zero-Shot Calibration}\\
            \midrule
             OOB & 7.32 & 13.34 & 19.07 \\
             TS & 2.02 & 6.74 & 11.81 \\
             LS & 3.2 & 6.93 & 12.1 \\
             TS + LS & 4.1 & 4.9 & 9.35\\
            \midrule
            \multicolumn{4}{c}{Few-Shot Calibration}\\
            \midrule
             Self-TS & 2.02 & 5.41 & 9.7 \\
             Self-TS + LS & 4.1 & 4.05 & 4.64 \\
             FSL & 7.32 & 7.67 & 9.23 \\
             FSL + LS & 3.2 & 4.37 & 5.73\\
             \bottomrule
        \end{tabular}
        \caption{Detailed results on XNLI with XLMR}
         \vspace*{3mm}
    \end{subtable}
    \hfill
    \begin{subtable}[h]{0.45\textwidth}
        \begin{tabular}{p{1.7cm}p{1cm}p{1.5cm}p{1.5cm}}
            \toprule
             Method & $\texttt{ECE}(en)$ & $\displaystyle \mathop{\mathbb{E}}_{l \in \mathcal{L}'}[\texttt{ECE}(l)]$ & $\displaystyle \max_{l \in \mathcal{L}'}\texttt{ECE}(l)$\\
            \midrule
            \multicolumn{4}{c}{Zero-Shot Calibration}\\
            \midrule
             OOB & 5.44 & 12.34 & 45.15 \\
             TS & 2.51 & 6.29 & 37.25 \\
             LS & 4.51 & 10.42 & 38.36 \\
             TS + LS & 2.61 & 6.71 & 30.51\\
            \midrule
            \multicolumn{4}{c}{Few-Shot Calibration}\\
            \midrule
             Self-TS & 2.51 & 4.77 & 29.82 \\
             Self-TS + LS & 2.61 & 4.7 & 23.23 \\
             FSL & 5.44 & 3.14 & 4.28 \\
             FSL + LS & 2.61 & 2.55 & 4.26\\
             \bottomrule
        \end{tabular}
        \caption{Detailed results on XNLI with mBERT}
         \vspace*{3mm}
    \end{subtable}
    
    \begin{subtable}[h]{0.45\textwidth}
        \begin{tabular}{p{1.7cm}p{1cm}p{1.5cm}p{1.5cm}}
            \toprule
             Method & $\texttt{ECE}(en)$ & $\displaystyle \mathop{\mathbb{E}}_{l \in \mathcal{L}'}[\texttt{ECE}(l)]$ & $\displaystyle \max_{l \in \mathcal{L}'}\texttt{ECE}(l)$\\
            \midrule
            \multicolumn{4}{c}{Zero-Shot Calibration}\\
            \midrule
             OOB & 14.54 & 20.01 & 29.33 \\
             TS & 12.31 & 15.95 & 24.04 \\
             LS & 9.66 & 5.47 & 9.42 \\
             TS + LS & 8.98 & 4.52 & 7.2\\
            \midrule
            \multicolumn{4}{c}{Few-Shot Calibration}\\
            \midrule
             Self-TS & 12.31 & 16.02 & 24.02 \\
             Self-TS + LS & 8.98 & 4.06 & 5.36 \\
             FSL & 14.54 & 8.93 & 14.92 \\
             FSL + LS & 9.66 & 4.4 & 5.74\\
             \bottomrule
        \end{tabular}
        \caption{Detailed results on XCOPA with XLMR}
         \vspace*{3mm}
    \end{subtable}
    \hfill
    \begin{subtable}[h]{0.45\textwidth}
        \begin{tabular}{p{1.7cm}p{1cm}p{1.5cm}p{1.5cm}}
            \toprule
             Method & $\texttt{ECE}(en)$ & $\displaystyle \mathop{\mathbb{E}}_{l \in \mathcal{L}'}[\texttt{ECE}(l)]$ & $\displaystyle \max_{l \in \mathcal{L}'}\texttt{ECE}(l)$\\
            \midrule
            \multicolumn{4}{c}{Zero-Shot Calibration}\\
            \midrule
             OOB & 23.4 & 23.51 & 29.02 \\
             TS & 21.76 & 20.02 & 23.99 \\
             LS & 15.85 & 12.41 & 15.87 \\
             TS + LS & 11.28 & 6.77 & 8.93\\
            \midrule
            \multicolumn{4}{c}{Few-Shot Calibration}\\
            \midrule
             Self-TS & 21.76 & 20.1 & 24.12 \\
             Self-TS + LS & 11.28 & 6.89 & 9.01 \\
             FSL & 23.4 & 3.75 & 10.5 \\
             FSL + LS & 15.85 & 3.54 & 6.65\\
             \bottomrule
        \end{tabular}
        \caption{Detailed results on XCOPA with mBERT}
         \vspace*{3mm}
    \end{subtable}

    \begin{subtable}[h]{0.45\textwidth}
        \begin{tabular}{p{1.7cm}p{1cm}p{1.5cm}p{1.5cm}}
            \toprule
             Method & $\texttt{ECE}(en)$ & $\displaystyle \mathop{\mathbb{E}}_{l \in \mathcal{L}'}[\texttt{ECE}(l)]$ & $\displaystyle \max_{l \in \mathcal{L}'}\texttt{ECE}(l)$\\
            \midrule
            \multicolumn{4}{c}{Zero-Shot Calibration}\\
            \midrule
             OOB & 7.15 & 9.65 & 13.45 \\
             TS & 2.75 & 4.22 & 5.8 \\
             LS & 5.36 & 7.93 & 10.66 \\
             TS + LS & 3.39 & 4.55 & 6.33\\
            \midrule
            \multicolumn{4}{c}{Few-Shot Calibration}\\
            \midrule
             Self-TS & 2.75 & 3.56 & 4.1 \\
             Self-TS + LS & 3.39 & 3.36 & 3.64 \\
             FSL & 7.15 & 2.51 & 3.51 \\
             FSL + LS & 5.36 & 2.59 & 3.28\\
             \bottomrule
        \end{tabular}
        \caption{Detailed results on MARC with XLMR}
         \vspace*{3mm}
    \end{subtable}
    \hfill
    \begin{subtable}[h]{0.45\textwidth}
        \begin{tabular}{p{1.7cm}p{1cm}p{1.5cm}p{1.5cm}}
            \toprule
             Method & $\texttt{ECE}(en)$ & $\displaystyle \mathop{\mathbb{E}}_{l \in \mathcal{L}'}[\texttt{ECE}(l)]$ & $\displaystyle \max_{l \in \mathcal{L}'}\texttt{ECE}(l)$\\
            \midrule
            \multicolumn{4}{c}{Zero-Shot Calibration}\\
            \midrule
             OOB & 9.38 & 11.1 & 17.3 \\
             TS & 3.56 & 6.14 & 10.9 \\
             LS & 5.19 & 5.96 & 12.9 \\
             TS + LS & 3.70 & 4.47 & 10.1\\
            \midrule
            \multicolumn{4}{c}{Few-Shot Calibration}\\
            \midrule
             Self-TS & 3.56 & 4.62 & 8.34 \\
             Self-TS + LS & 3.70 & 4.71 & 7.50 \\
             FSL & 9.38 & 2.12 & 3.45 \\
             FSL + LS & 5.19 & 3.02 & 4.24\\
             \bottomrule
        \end{tabular}
        \caption{Detailed results on MARC with mBERT}
         \vspace*{3mm}
    \end{subtable}

    \begin{subtable}[h]{0.45\textwidth}
        \begin{tabular}{p{1.7cm}p{1cm}p{1.5cm}p{1.5cm}}
            \toprule
             Method & $\texttt{ECE}(en)$ & $\displaystyle \mathop{\mathbb{E}}_{l \in \mathcal{L}'}[\texttt{ECE}(l)]$ & $\displaystyle \max_{l \in \mathcal{L}'}\texttt{ECE}(l)$\\
            \midrule
            \multicolumn{4}{c}{Zero-Shot Calibration}\\
            \midrule
             OOB & 1.93 & 4.28 & 5.88 \\
             TS & 0.81 & 2.29 & 3.39 \\
             LS & 4.83 & 3.37 & 3.85 \\
             TS + LS & 6.30 & 4.44 & 5.07\\
             \bottomrule
        \end{tabular}
        \caption{Detailed results on PAWS-X with XLMR}
         \vspace*{3mm}
    \end{subtable}
    \hfill
    \begin{subtable}[h]{0.45\textwidth}
        \begin{tabular}{p{1.7cm}p{1cm}p{1.5cm}p{1.5cm}}
            \toprule
             Method & $\texttt{ECE}(en)$ & $\displaystyle \mathop{\mathbb{E}}_{l \in \mathcal{L}'}[\texttt{ECE}(l)]$ & $\displaystyle \max_{l \in \mathcal{L}'}\texttt{ECE}(l)$\\
            \midrule
            \multicolumn{4}{c}{Zero-Shot Calibration}\\
            \midrule
             OOB & 3.57 & 10.3 & 15.6 \\
             TS & 0.99 & 5.64 & 9.80 \\
             LS & 3.35 & 5.57 & 8.94 \\
             TS + LS & 5.27 & 5.36 & 7.40\\
             \bottomrule
        \end{tabular}
        \caption{Detailed results on PAWS-X with mBERT}
         \vspace*{3mm}
    \end{subtable}
\caption{Detailed results on improving calibration for the 4 datasets and 2 MMLMs considered in our experiments}
\label{tab:det_results}
\end{table*}

% \begin{figure*}
%     \centering
%     % \captionsetup[subfloat]{margin=0.5em}
%     \begin{subfigure}{.2\textwidth}
%     \centering
%     \captionsetup{justification=centering}
%     \includegraphics[width=.99\textwidth]{figures/en_oob.pdf}
%     \vspace*{-7mm}
%     \caption{}%Out of box calibration on English}
%     \label{fig:en_oob_full}
%     \end{subfigure}
%     \begin{subfigure}{.2\textwidth}
%     \centering
%     \includegraphics[width=.99\textwidth]{figures/sw_oob.pdf}
%     \vspace*{-7mm}
%     \caption{}%Out of box calibration on Swahili}
%     \label{fig:sw_oob_full}
%     \end{subfigure}
%     \begin{subfigure}{.2\textwidth}
%     \centering
%     \includegraphics[width=.99\textwidth]{figures/sw_zs_cali.pdf}
%     \vspace*{-7mm}
%     \caption{}%Calibrating using zero-shot methods}
%     \label{fig:sw_zs_full}
%     \end{subfigure}
%     \begin{subfigure}{.2\textwidth}
%     \centering
%     \includegraphics[width=.99\textwidth]{figures/sw_fs_cali.pdf}
%     \vspace*{-7mm}
%     \caption{}%Calibrating using few-shot methods}
%     \label{fig:sw_fs_full}
%     \end{subfigure}
%     \vspace*{-3mm}
%     \caption{Full version of the reliability diagrams shown in Figure \ref{fig:cal_en_sw} also containing a comparison between the average confidence and accuracy of the model.}
%     \vspace*{-3mm}
%     \label{fig:cal_en_sw_det}
% \end{figure*}

\begin{figure*}
    \centering
    \includegraphics[width=.9\textwidth]{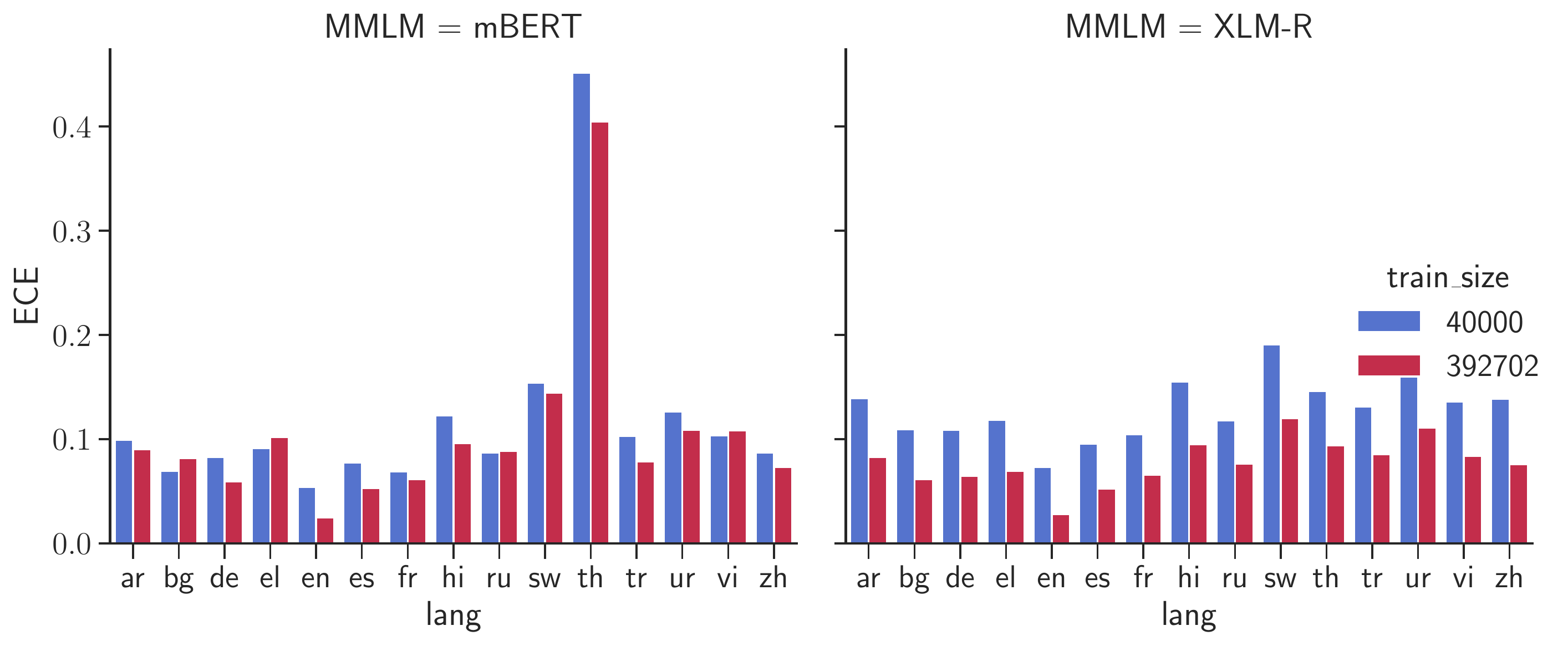}
    \caption{Out-of-Box Calibration Errors (ECE) for XLMR and mBERT trained on XNLI with 40k samples (sub-sampled) and 392k samples(full data). Even though the calibration is better with using the entire data for training, the observed patterns about the models being mis-calibrated on languages other than English, especially low resource languages like Swahili, Thai and Urdu still hold true.}
    \label{fig:full_v_sample}
\end{figure*}

\begin{figure}
    \centering
    \begin{subfigure}{.49\textwidth}
        \includegraphics[width=.99\textwidth]{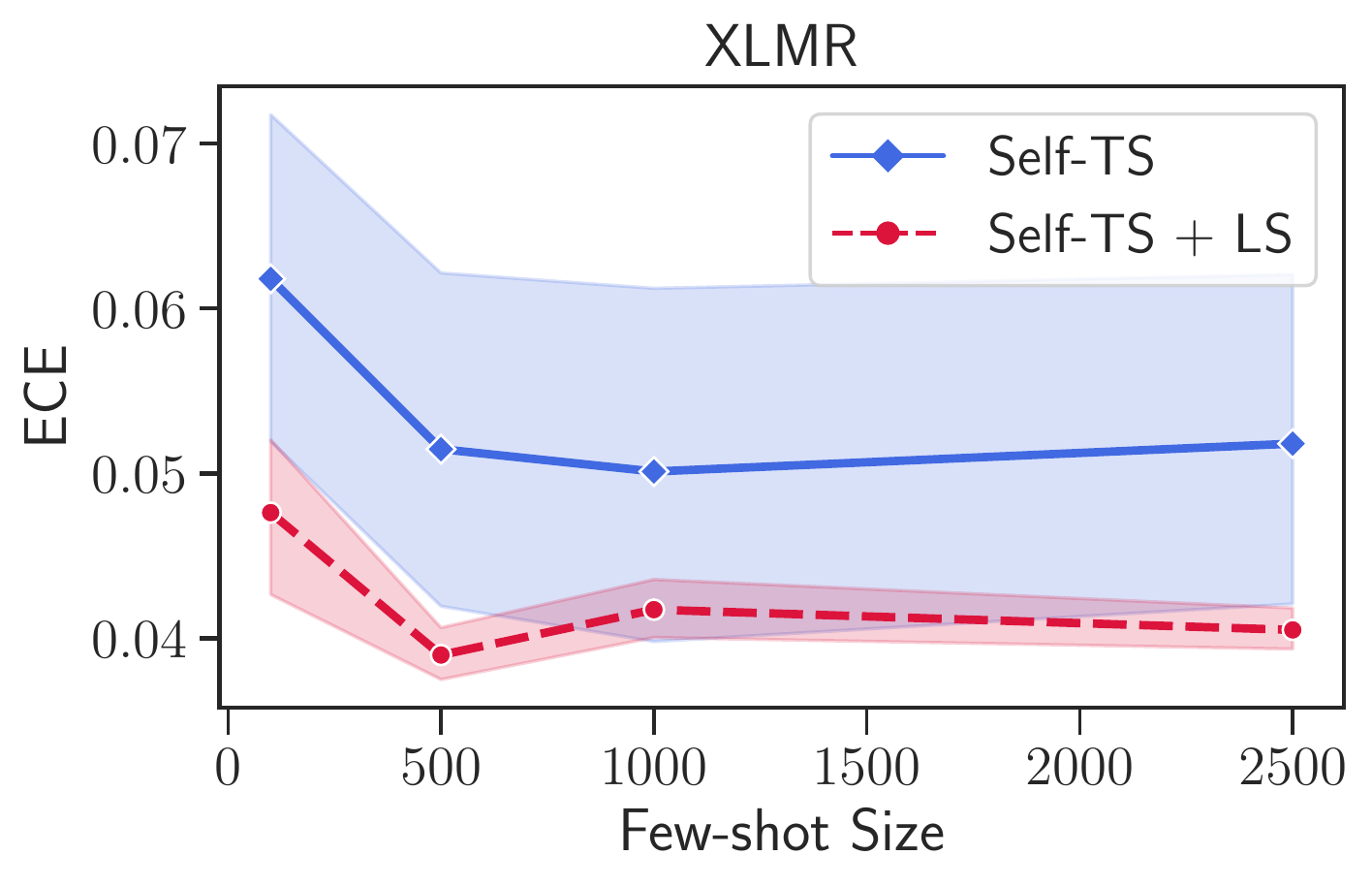}
    \caption{}
    \label{fig:xlmr_size_vary}
    \end{subfigure}
    \begin{subfigure}{.49\textwidth}
        \includegraphics[width=.99\textwidth]{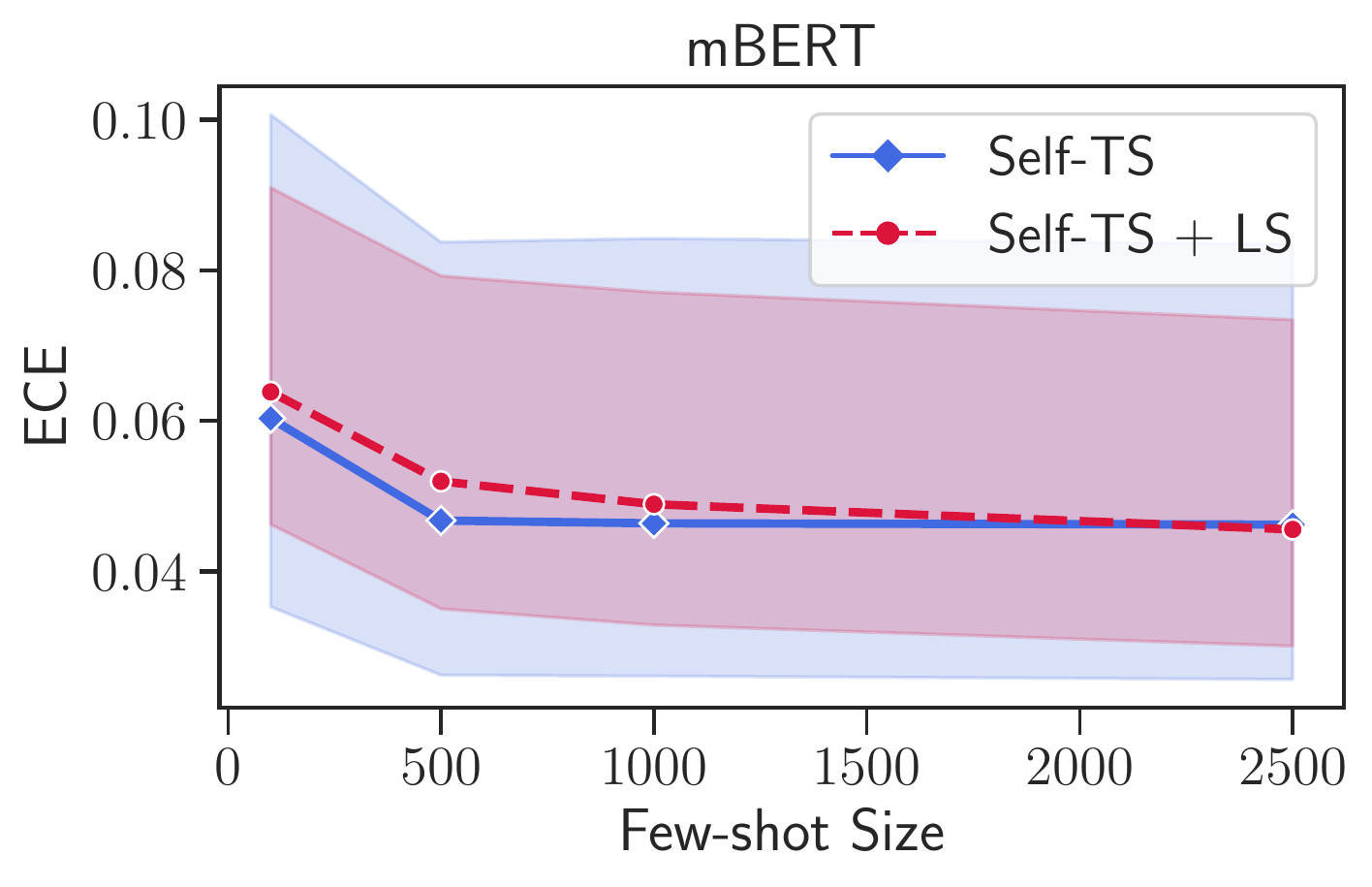}
    \caption{}
    \label{fig:mbert_size_vary}
    \end{subfigure}
\caption{Variation in ECE as we use more and more data for calibrating using Self-TS method across languages for XNLI dataset. As can be seen 500 samples are sufficient to obtain low calibration errors.}
\end{figure}

\begin{figure}
    \centering
    \begin{subfigure}{.49\textwidth}
        \includegraphics[width=.99\textwidth]{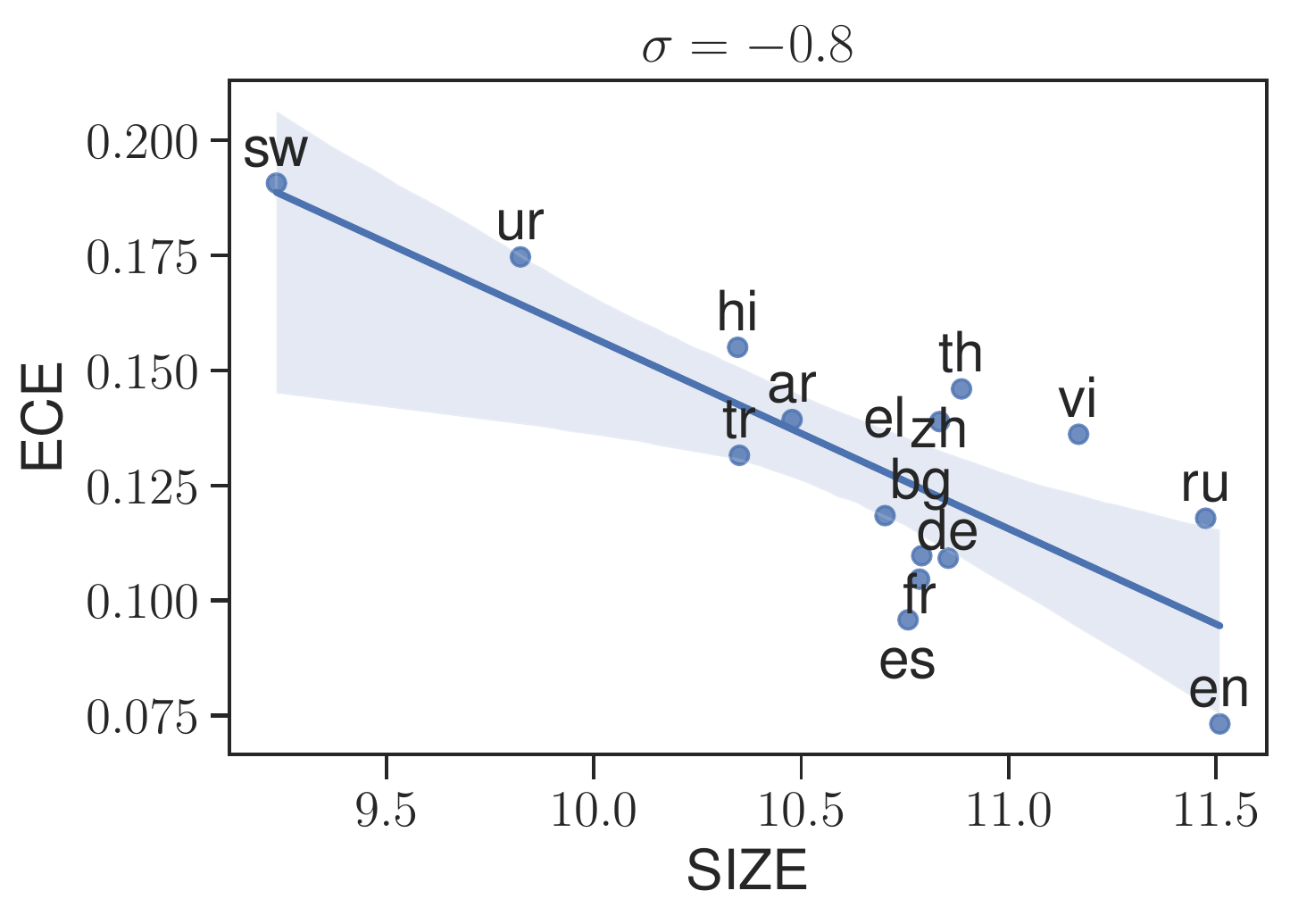}
    \caption{}
    \label{fig:size_corr}
    \end{subfigure}
    \begin{subfigure}{.49\textwidth}
        \includegraphics[width=.99\textwidth]{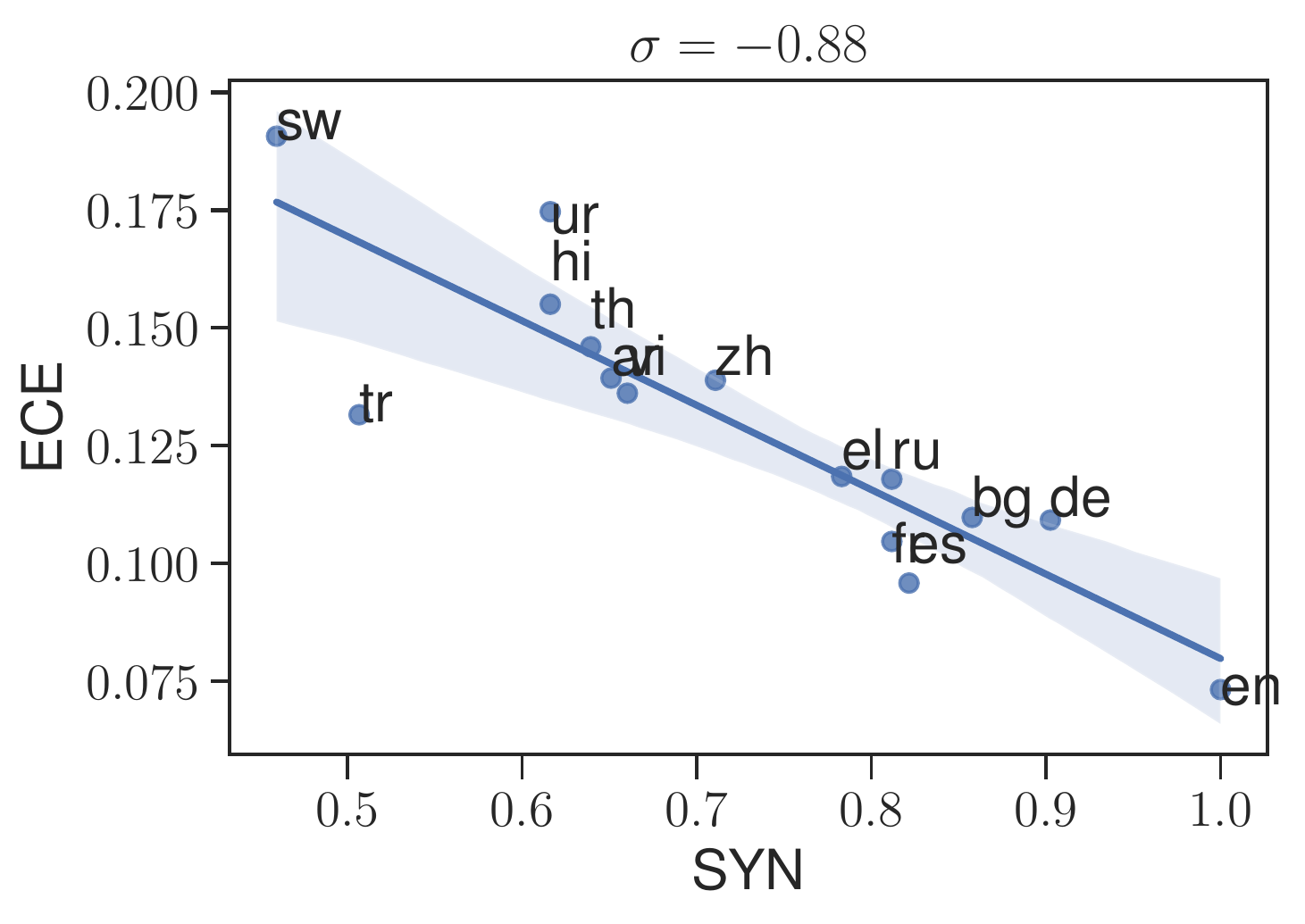}
    \caption{}
    \label{fig:syn_corr}
    \end{subfigure}
    \begin{subfigure}{.49\textwidth}
        \includegraphics[width=.99\textwidth]{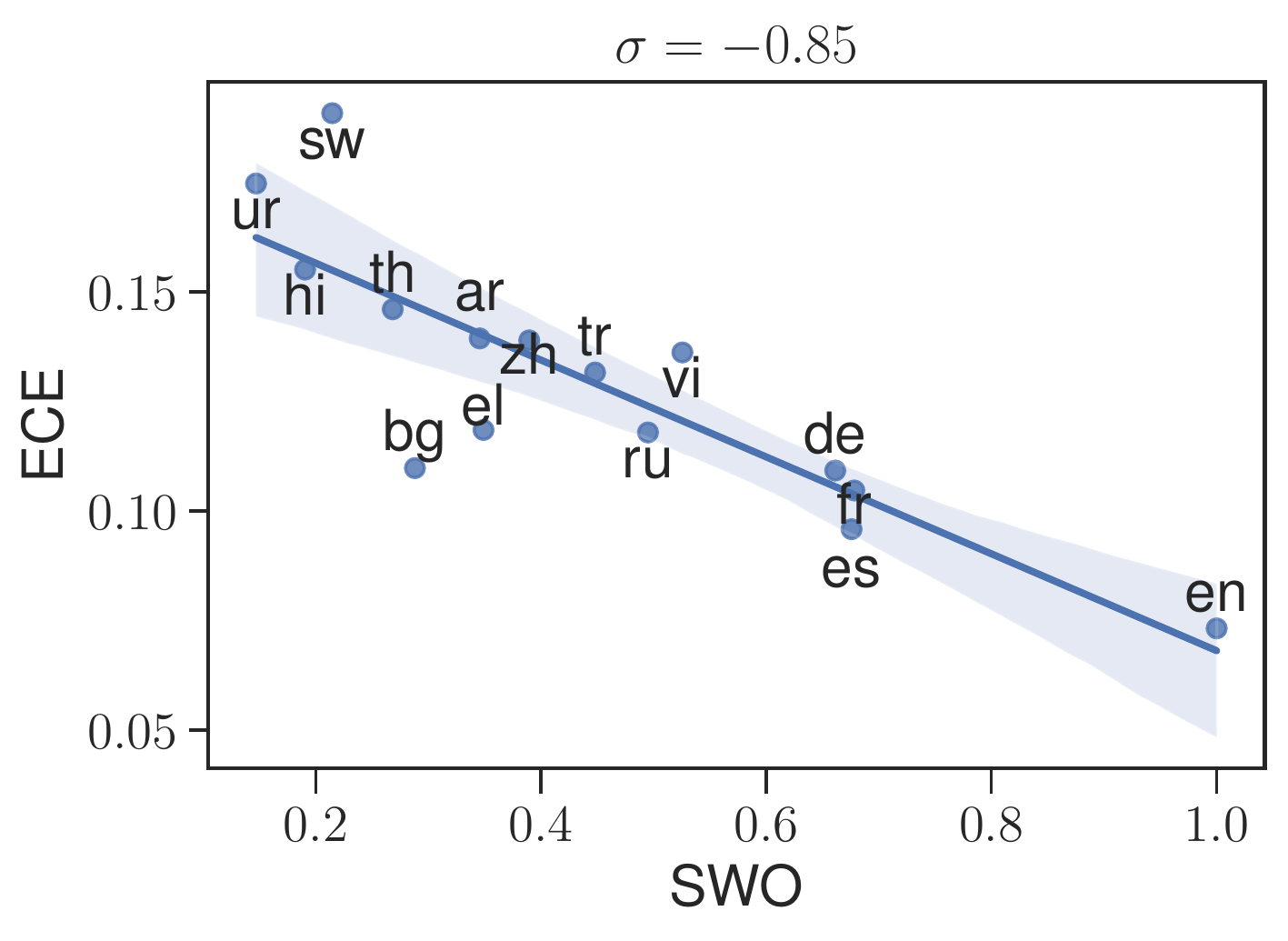}
    \caption{}
    \label{fig:swo_corr}
    \end{subfigure}
\caption{Visualizing the correlations of ECE with SIZE, SYN and SWO for XLMR fine-tuned on XNLI.}
\end{figure}

\end{document}